\documentclass[10pt,twocolumn,letterpaper]{article}

\usepackage{wacv}
\usepackage{times}
\usepackage{epsfig}
\usepackage{graphicx}
\usepackage{amsmath}
\usepackage{amssymb}

\usepackage{subcaption}
\usepackage{xcolor}
\usepackage{stix}
\usepackage{makecell}
\usepackage{pdfpages}
\usepackage{enumitem}

\newcommand{\ngf}{{\textit{ngf}}}

\usepackage[pagebackref=true,breaklinks=true,letterpaper=true,colorlinks,bookmarks=false]{hyperref}

\usepackage[pagebackref=true,breaklinks=true,letterpaper=true,colorlinks,bookmarks=false]{hyperref}
\wacvfinalcopy 

\def\wacvPaperID{254} %

\begin{document}

\title{TwoStreamVAN: Improving Motion Modeling in Video Generation}

\author{Ximeng Sun \footnotemark[2]
\hspace{10mm} Huijuan Xu \footnotemark[2] \hspace{2pt} \footnotemark[3] \hspace{10mm}
Kate Saenko \footnotemark[2] \hspace{10mm}
\\ 
\footnotemark[2] Boston University
\hspace{10mm}
\footnotemark[3] University of California, Berkeley \\
\footnotemark[2] {\tt\small \{sunxm, hxu, saenko\}@bu.edu} \hspace{6mm} \footnotemark[3] {\tt\small huijuan@berkeley.edu}
}

\maketitle
\thispagestyle{empty}
\begin{abstract}
Video generation is an inherently challenging task, as it requires modeling realistic temporal dynamics as well as spatial content. Existing methods entangle the two intrinsically different tasks of motion and content creation in a single generator network, but this approach struggles to simultaneously generate plausible motion and content. To improve motion modeling in video generation task, we propose a two-stream model that disentangles motion generation from content generation, called a Two-Stream Variational Adversarial Network (TwoStreamVAN). Given an action label and a noise vector, our model is able to create clear and consistent motion, and thus yields photorealistic videos. 
The key idea is to progressively generate and fuse multi-scale motion with its corresponding spatial content.
Our model significantly outperforms existing methods on the standard Weizmann Human Action, MUG Facial Expression and VoxCeleb datasets, as well as our new dataset of diverse human actions with challenging and complex motion. Our code is available at \url{https://github.com/sunxm2357/TwoStreamVAN/}.

\end{abstract}

 \vspace{-10pt}
 \section{Introduction}
Despite great progress being made in generation of well-structured static images~\cite{liu2015faceattributes, yu2015lsun} using methods such as GANs/VAEs (\cite{chen2016infogan, hong2018inferring, odena2016conditional, radford2015unsupervised, reed2016generative, mao2017least, kodali2017convergence, karras2017progressive, karras2018style}), generation of pixel-level video has yet to achieve similarly impressive results~\cite{vondrick2016generating, saito2017temporal, tulyakov2017mocogan, he2018probabilistic}. The challenge of video generation lies in the need to construct unstructured spatial content with both foreground objects and background, and simultaneously model natural motion at different scales and locations. 
Existing methods fail to generate convincing videos (see examples in Fig.~\ref{fig:comparison_1st_page} 
or the video in the supplementary material) mostly because of ineffective motion modeling,
which in turn deteriorates content generation. A major problem with pixel-level video prediction~\cite{mathieu2015deep, srivastava2015unsupervised, villegas2017decomposing} and generation methods is that they attempt to model both static content and dynamic motion in a single entangled generator, regardless of whether they disentangle the motion and content in the latent space or not.

\begin{figure}
    \begin{subfigure}[b]{\linewidth}
		\centering
        \includegraphics[width=\linewidth]{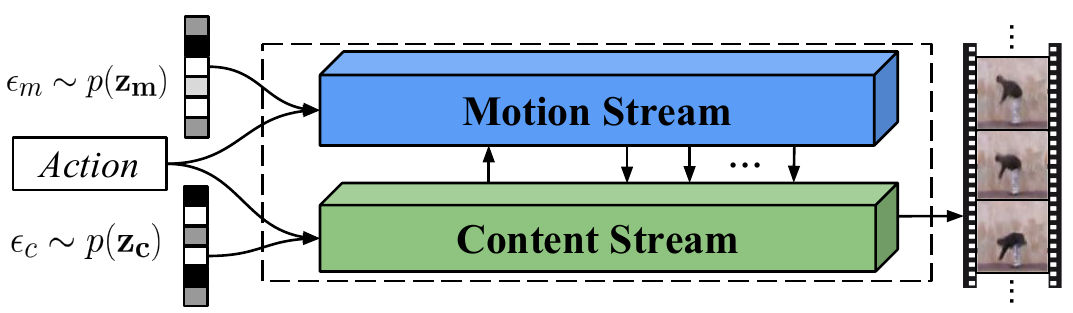}
        \caption{Overview of Our \textit{TwoStreamVAN}}
        \label{fig:problem_statement}
	\end{subfigure} \hfill
	\begin{subfigure}[b]{\linewidth}
		\centering
        \includegraphics[width=\linewidth]{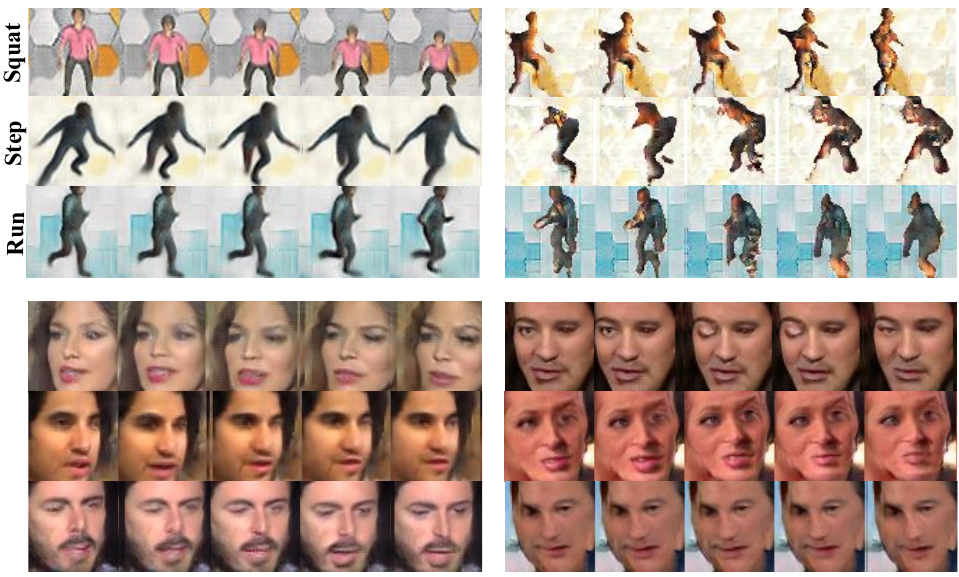}
        \caption{TwoStreamVAN (left) v.s. MoCoGAN (right)}
        \label{fig:comparison_1st_page}
	\end{subfigure}
	\caption{\small We propose Two-Stream Variational Adversarial Network (TwoStreamVAN) to improve motion modeling. Given an action class, it separately generates content and motion from the disentangled content and motion latent codes $\epsilon_c$ and $e_m$ and fuse them via a novel multi-scale fusion mechanism. Our model generates significantly better videos than MoCoGAN without separation motion generation.}
	 \vspace{-15pt}
\end{figure}

We argue that separating motion and content modeling in the decoding phase is crucially important. On one hand, it removes the disturbance of motion during  content generation and results in better content structures.  On the other hand,  separate motion modeling helps to produce action-relevant movement consistently across the entire sequence.

We propose a novel \textit{Two-Stream Variational Adversarial Network (TwoStreamVAN)} that generates a video from an action label and a disentangled noise vector
(Fig.~\ref{fig:problem_statement})
. Rather than over-estimating a single generator's ability, we introduce two parallel generators that process content and motion separately and fuse them together to predict the next frame. Intuitively, motion is usually conditioned on the spatial information within its local context window, e.g., different motions near the arms and legs when doing jumping jacks. Thus, we define the fusion of motion and content  as a learned refinement of pixel values unique to each location. 
To precisely generate multi-scale motion, we conduct such refinement on intermediate content layers with several different resolutions.
Moreover, we introduce motion mask to make the learning of motion only focus on the regions where it exists.

Considering the great success achieved in image generation, we expect that introducing  image-level supervision will reduce the  deterioration of content generation common in most video generation models. Previously, MoCoGAN~\cite{tulyakov2017mocogan} tried image-level supervision but made little progress, since the content generation was still affected by the motion modeling in a single shared generator. Notably, a key advantage of our disentangled two-stream generators is the ability to learn each stream separately and thus more accurately. We learn the image structure in its own generator fully via image-level supervision, which significantly enhances content generation performance. Well-trained content generation further benefits motion learning in video-level supervision.

We evaluate our approach on two standard video generation benchmarks, Weizmann Human Action~\cite{ActionsAsSpaceTimeShapes_pami07} and MUG Facial Expression~\cite{aifanti2010mug} as well as a large-scale speech video dataset -- VoxCeleb~\cite{Nagrani17}. 
To verify the advantage of our two-stream generator in modeling motion, we propose a large synthetic human action dataset, called \textit{SynAction}, with  challenging motion complexity (120 unique action models across 20 different non-rigid human actions, like running or squatting), 
using a library of video game actions, Mixamo~\cite{corazza2014real}.
To summarize,  our contributions are:
\begin{itemize}[leftmargin=*]
    \itemsep0em
    \item We propose a video generation model \textit{TwoStreamVAN} as well as a more effective learning scheme, which disentangle motion and content in the generation phase. 
    \item We design a multi-scale motion fusion mechanism and further
    improve motion modeling by conditioning on the spatial context;
    \item We create a large-scale synthetic video generation dataset available to the research community; 
    \item We evaluate our model on four video datasets both quantitatively and qualitatively (with user studies), and demonstrate superior results to several strong baselines.
\end{itemize}


\section{Related Work}
\textbf{Generative Models.} The two main deep learning methods for image generation are VAEs~\cite{kingma2013auto, gulrajani2016pixelvae, gregor2015draw, yan2016attribute2image} and GANs~\cite{goodfellow2014generative, odena2016conditional, radford2015unsupervised, chen2016infogan}. VAEs provide probabilistic descriptions of observations in latent spaces, but  might generate blurry and unrealistic images in their vanilla form. GANs propose an adversarial discriminator to encourage the generation of crisper images, but suffer from mode collapse~\cite{zhou2017activation} and unexpected bizarre artifacts. \cite{larsen2015autoencoding, makhzani2015adversarial} combine a VAE and a GAN and propose a Variational Adversarial Network (VAN) to learn an interpretable latent space as well as generate realistic images. In the light of the VAN's success in image generation, we adopt it here for video generation. 

\textbf{Video Generation} is a challenging video task, which maps random noise to the content and motion which form a plausible video sequence.
VideoVAE~\cite{he2018probabilistic} shows the VAE's ability to produce video, proposing a structured latent space and an encoder-generator architecture to generate the video recurrently. Except for this work, the prevailing trend is to generate videos from noise utilizing the GAN training paradigm. VideoGAN~\cite{vondrick2016generating} generates a sequence of foreground objects with a single shared static background. Instead of generating foreground and background separately, TGAN~\cite{saito2017temporal} decodes each frame from a unified spatio-temporal latent representation. MoCoGAN attempts to disentangle content and motion by sampling from separate latent spaces but uses a single generator to decode these two latent codes together. To overcome the ineffective motion modeling and the consequent content deterioration in the unified generation process, we further introduce disentangled content and motion generators to model spatial structures and temporal dynamics separately.

\textbf{Video Prediction}~\cite{denton2017unsupervised, finn2016unsupervised, denton2018stochastic, villegas2017decomposing, babaeizadeh2017stochastic, lee2018stochastic, srivastava2015unsupervised}, a related yet different task, outputs additional frames for a partial input video by borrowing the content and extrapolating the motion from the given input video.  The main difference of two tasks lies in the origins of the content and motion in the new output sequence: the generation samples them from noise, but the prediction borrows the content and extrapolates the motion from the input frames. Thus, they adopt different metrics, emphasis on the future reconstruction and the generation reality/diversity respectively. Despite of differences, they share some technical details in common. For instance, \cite{villegas2017decomposing, denton2017unsupervised} decompose the content and motion in the encode phase; \cite{finn2016unsupervised} discusses different motion transformations as effective motion representations. In this paper, we mainly focus on Video Generation and show our superiority over these shared techniques in the pure generative context.

\textbf{Multi-scale Motion Estimation and Prediction.} To tackle multi-scale motion in actual videos, \cite{black1996robust, brox2004high, sun2014quantitative, ranjan2017optical} build a pyramid of real images, while \cite{sun2018pwc, xue2016visual} constructs a pyramid of embedded feature maps. In our model, we take the second method. Traditionally motion between two adjacent frames is introduced by warping the dense optical flow features with the current frame \cite{krishnamurthy1999frame, kaviani2016frame}. To avoid the high computational cost of optical flow, Xue~\etal~\cite{xue2016visual} generates multi-scale motion kernels from the difference map of adjacent frames. 
They learn generic motion kernels shared for the entire image which is hard to interpret, because motion is usually conditioned on the local spatial context. 
In this paper, we produce motion kernels specifically for each spatial location and use these kernels to refine intermediate layers in content feature pyramid to model multi-scale motion. Furthermore, as \cite{xue2016visual} only predicts the next frame from the current, we design our model to address a more challenging task which generates the video sequence without receiving any visual clue as inputs.


\begin{figure*}
\begin{center}
     \includegraphics[width=\linewidth]{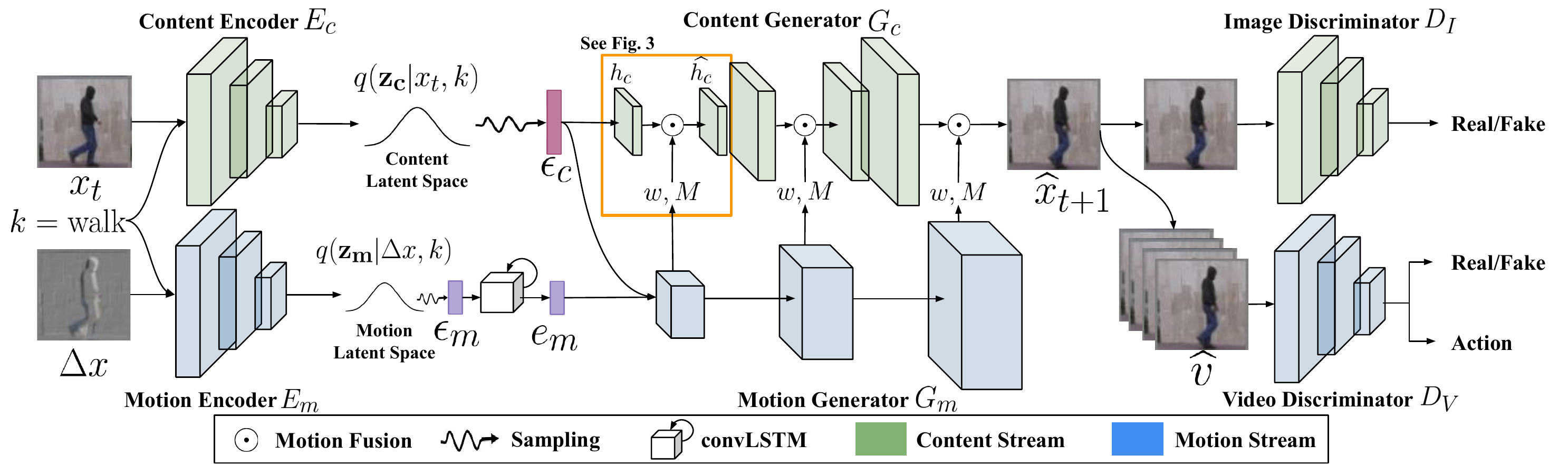}
     \vspace{-25pt}
\end{center}
   \caption{\small Our Two-Stream Variational Adversarial Network learns to generate the next frame $\widehat x_{t+1}$ from the current frame $x_t$. At training time, Content and Motion Encoders ($E_c$ and $E_m$) generate latent distributions $q \left (\mathbf{z_c}|x_t, k \right )$ and $q \left(\mathbf{z_m}|\Delta x, k \right)$ by viewing $x_t$ and the difference map $\Delta x$ between $x_{t+1}$ and $x_t$ respectively. From latent distributions, we sample the content latent vector $\epsilon_c$ and $\epsilon_m$ and employ a convLSTM to get $e_m$ encoding the history of temporal dynamics. 
   Content and Motion Generators ($G_c$ and $G_m$) take in $\epsilon_c$ and $e_m$ and then decode a content hidden layer $h_c$ and motion kernels $w$ along with a motion mask $M$ at each scale. $h_c$, $w$ and $M$ are served as inputs to our defined multi-scale motion fusion (Fig.~\ref{fig:motion_refine}).
   Image and Video Discriminators ($D_I$ and $D_V$) encourage the model to generate both realistic content and motion.}
   \vspace{-10pt}
   \label{fig:model_overview}
\end{figure*}

\section{Approach}
We introduce a Two-Stream Variational Adversarial Network (Fig.~\ref{fig:model_overview}), which generates a video given an input action label and a random noise vector.
We define action-conditioned video generation as follows. Suppose we have $K$ different action classes. In each class $k \in 1\text{:} K$, let $C_k$ be the number of training videos. Let $V_k= \left \{v_{i, k}, \forall i \in 1 \text{:} C_k \right \}$ be the set of all  videos for class $k$ and  $v_{i, k}= \left \{x_1, x_2, \cdots, x_T \right \}$ be a short video clip with $T$ frames.
The task is to define a function $G$ which generates a plausible video $\widehat v = \left \{\widehat{x}_1, \cdots, \widehat{x}_T \right \}$ conditioned on the given class label $k$ from a latent vector $\epsilon \in \mathbb{R} ^ \mathcal{N}$, i.e. $\widehat v = G \left(k, \epsilon \right)$.

We disentangle the latent space $\epsilon$ into two independent codes: a content code $\epsilon_c \in \mathbb{R}^ \mathcal{C} $, and a motion code $\epsilon_m \in \mathbb{R}^ \mathcal{M}$, with $\mathcal{N}= \mathcal{C}+ \mathcal{M}$. We also disentangle the video generation function $G$ into two separate content and motion  functions ($G_c$ and $G_m$), 
in contrast to previous work that used a single generator~\cite{vondrick2016generating, saito2017temporal, tulyakov2017mocogan}, and we design a novel multi-scale fusion mechanism.

\subsection{Two-Stream Generation}
To learn generative functions $G_c$ and $G_m$ for content and motion modeling, we introduce two separate action-conditioned VAN streams with interactions at several stages. Each VAN stream contains an encoder, a generator and a discriminator, where the encoder and generator serve as the auto-encoder in the VAE, and the generator and discriminator comprise the GAN.

The \textbf{Content VAN Stream} consists of a Content Encoder $E_{c}$, a Content Generator $G_{c}$ and an Image Discriminator $D_{I}$. 
After observing a single frame $x$, $E_{c}$ generates the posterior content latent distribution $q \left (\mathbf{z_c} |x, k \right )$, which is close to its true prior distribution $p \left (\mathbf{z_c} |k \right )$. 
$G_{c}$ decodes a content vector $\epsilon_c$ sampled from the content distribution into a frame $\widehat{x}$. $D_{I}$ discriminates  real/generated frames to encourage $G_c$ to generate realistic image patterns.

Similarly, the \textbf{Motion VAN Stream} consists of a  Motion Encoder $E_m$, a Motion Generator $G_m$ and a Video Discriminator $D_V$. 
Instead of encoding spatial content, in our approach $E_m$ models the temporal dynamics in the difference map $\Delta x$ between neighbor frames. It generates the posterior motion latent distribution
$q \left(\mathbf{z_m} | \Delta x, k \right)$ which is close to its true prior distribution $p \left (\mathbf{z_m} |k \right)$. 
A convLSTM~\cite{xingjian2015convolutional} accumulates the motion history and generates the current motion embedding $e_m$ by receiving a sequence of $\epsilon _m$ sampled from the motion distributions at all previous time steps. $G_m$ takes in $\epsilon_c$ and $e_m$ to generate motion at different scales. We generate each video $\widehat{v}$ ~by fusing the generated motion with the corresponding content at $T$ time steps
(see Sec.~\ref{sec:motion_refinement}).
$D_V$ discriminates the real/generated videos and additionally classifies their actions to encourage $G_m$ to generate  realistic motion for action $k$.

\subsection{Multi-scale Motion Generation and Fusion} \label{sec:motion_refinement}

At pixel $\left (a,b \right)$, the motion usually happens within a local window between adjacent frames. Inspired by the spatial convolution for frame interpolation~\cite{niklaus2017video},
we represent motion as a refinement of the current pixel value based on its local context and fuse such motion with content via spatially-adaptive convolution.
Moreover, we propose a novel multi-scale fusion mechanism to overcome the drawbacks of their approach, namely: 1) ineffective modeling of multi-scale motion due to the single fusion step performed on the full-resolution image and 2) the high demand for memory due to the large convolution kernels used to represent the maximum possible motion. 

\begin{figure}
    \begin{center}
         \includegraphics[width=\linewidth]{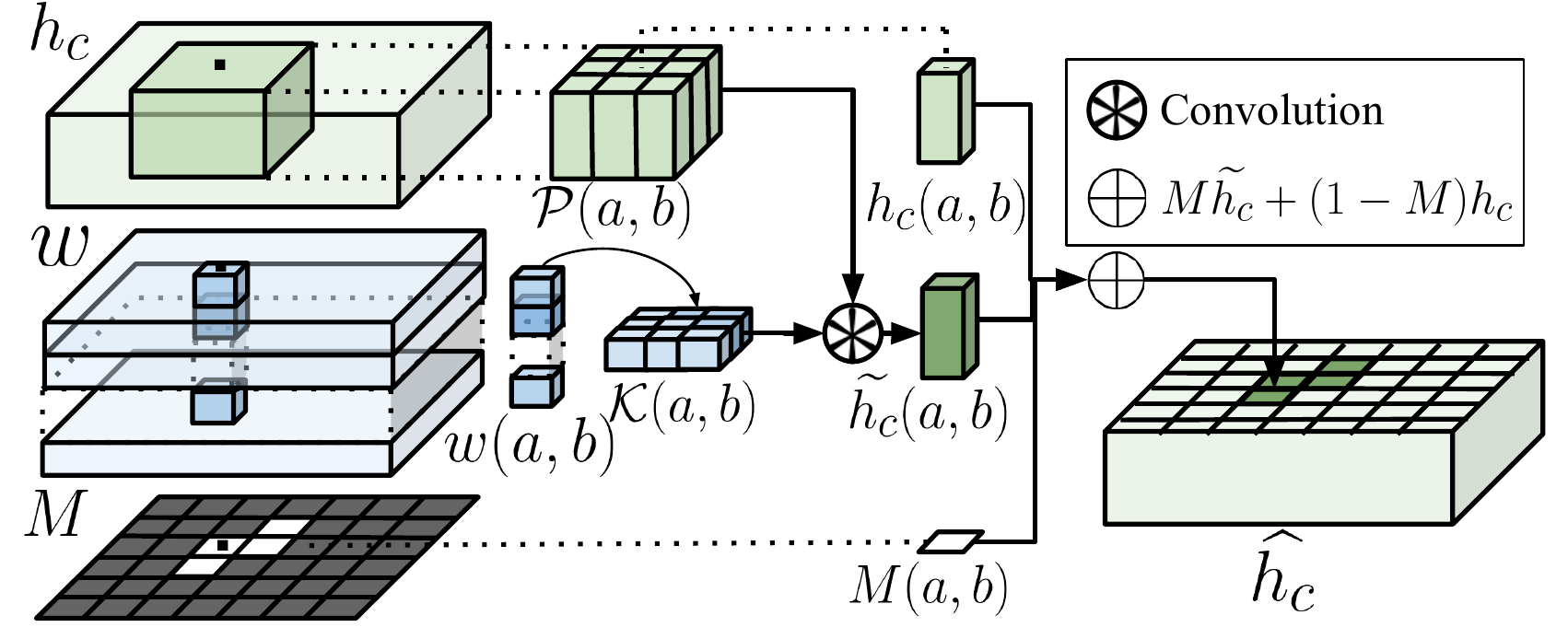}
    \vspace{-25pt}
    \end{center}
   \caption{\small Motion fusion for location $\left(a,b \right)$ on its local window $\mathcal{P} \left(a,b \right)$, at scale $s$. Take kernel size $n=3$ for example. $h_c$ is the content hidden layer from $G_c$. $w$ and $M$ are  convolutional kernels and the motion mask from $G_m$. We first recover the 2D kernel $\mathcal{K} \left (a, b \right )$ from its flattened form $w \left(a, b \right)$. Then we compute an intermediate content feature $\widetilde{h_c} \left(a,b \right)$ by convolving 
   $\mathcal{P} \left(a,b \right)$ with $\mathcal{K} \left(a,b \right)$. Finally, we update $h_c$ with $\widetilde{h_c}$ guided by $M$ to get the new hidden layer $\widehat{h}_c$.}
   \vspace{-15pt}
   \label{fig:motion_refine}
\end{figure}

To generate  precise motion, $G_m$ takes the disentangled content and motion embeddings ($\epsilon_c$ and $e_m$) as input.
In $G_m$, motion at different scales is separated at the corresponding hidden layers:
large motion comes from low resolution layers and small motion from high resolution layers. 
At each layer, $G_m$ uses the current feature map to: 1) calculate the motion (pixel refinement) in the form of pixel-wise 2D kernels with size $n$; 2) identify the regions containing actual movements in a motion mask; 3) generate the motion map for the following layer.
To fuse motion with content, we convolve the generated 2D kernels with patches centered at their corresponding pixels respectively (i.e. perform spatially-adaptive convolution) only if the pixel is in the regions where motion is activated (see Fig.~\ref{fig:motion_refine}).

Specifically, suppose we separate the motion into $S$ scales. For each scale $s$, let $l_s$ be the resolution of its corresponding hidden layer, $w^s \in \mathbb{R} ^ {l_s \times l_s \times n^2 }$ be the convolution kernels produced by $G_m$ and  $h_c^s \in \mathbb{R}^{l_s \times l_s \times d_s} $ be the corresponding content layer, where $d_s$ is the content feature dimension.  
We perform the spatially-adaptive convolution in the following steps.
First, for each location $(a,b)$, we recover a 2D convolution kernel $\mathcal{K}^s \left(a, b \right) \in \mathbb{R}^{n \times n}$ from its flattened form $w^s \left(a,b \right)$. Then, we convolve $\mathcal{K}^s \left(a, b \right)$ with the content local patch $\mathcal{P}^s \left(a,b \right) \in \mathbb{R}^{n \times n \times d_s} $ on $h_c^s$ to produce an intermediate content representation $\widetilde{h^s_c} \left(a, b \right)$ for the location $\left(a, b \right) $:
\small
\begin{align} 
    \widetilde{h^s_c} \left(a, b \right) =  \mathcal{K}^s \left(a, b\right) * \mathcal{P}^s \left(a,b \right).
\end{align}
\normalsize

In Sec~\ref{sec:multi-scale_ablation}, we show the effectiveness of our approach for capturing motion at different scales. Moreover, we handle larger motion by adding adaptive convolutions to smaller layers while \cite{niklaus2017video} handles it by increasing the kernel size. Since the number of parameters to store and model is quadratic in the kernel size, we significantly reduce the memory usage and the model complexity by leveraging small kernels ($n=3$ or 5) for all layers in our multi-scale fusion framework.

To focus $G_m$'s attention on learning the motion of scale $s$ in the regions where it actually happens, we predict a motion mask $M^s \in \mathbb{R}^{l_s \times l_s}$ along with $w^s$, to identify such regions. Each entry $M^s \left(a,b \right)$ is in $ \left[0, 1 \right]$. We generate the new content map $\widehat{h_c^s}$ with fused motion from  $h_c^s \left(a, b\right)$ and $\widetilde {h^s_c} \left(a, b\right)$ guided by $M^s(a, b)$:
\small
\begin{align}
    \widehat{h_c^s} \left(a, b \right) =  M^s \left(a,b \right) \widetilde{h_c^s} \left(a, b \right)  + \left(1- M^s \left(a,b \right)\right) h_c^s \left(a, b\right).
\end{align}
\normalsize

In addition to focusing the model's attention, motion mask helps preserve the pixel value in the refinement by simply deactivating $M^s(a, b)$. In contrast, \cite{niklaus2017video} requires $G_m$ to learn a 2D kernel $\mathcal{K}$ with a special pattern for pixel preservation. In Sec.~\ref{sec:mask_visual}, we show that our masks activate the correct areas at different scales and achieve better motion generation and background preservation.

\subsection{Learning}

Taking advantage of our disentangled content and motion streams, we propose an alternating dual-task learning scheme to learn each stream separately and thus more accurately and effectively. Specifically, the content stream is learned via image reconstruction, while the motion stream is learned via video reconstruction. We alternate training, such that each stream is trained while the other is fixed, and use both VAE and GAN losses to optimize each stream.

\vspace{.1in}\noindent\textbf{Content Learning.}
We learn our content stream solely by reconstructing the current frame $x$, without the disturbance of motion modeling, and reach  compatible performance with image generation methods.
We optimize the content VAE loss and image-level GAN loss. The VAE loss includes a reconstruction loss for the current frame and a KL-divergence between the prior content latent distribution $ p\left(\mathbf{z_c} |k\right)$ and the posterior distribution $q \left(\mathbf{z_c} | x, k \right)$:
\small{
\begin{align}
   \min\limits_{E_c, G_c}  \mathcal{L}_{VAE-{c}} = &\lambda_1 \mathcal{L}_2 \left(\widehat{x}, x\right) + \lambda_{2} KL \left(q \left(\mathbf{z_c} | x, k \right)|| p \left(\mathbf{z_c} |k\right) \right).
\end{align}
}
\normalsize
Image-level GAN training uses  real frames $x$ as  positive examples, and treats the generated images $x_p$ and $\widehat{x}$ sampled from $p \left(\mathbf{z_c} | k \right)$ and $q \left(\mathbf{z_c} | x, k \right)$ as  negative examples following Larsen \etal \cite{larsen2015autoencoding}:
\small
\begin{align}     
     \max\limits_{G_c} \min\limits_{D_I} \mathcal{L}_{GAN-c} = & \ \text{log} \left[1- D_{I} \left( x \right) \right] + \text{log}\left[D_{I} \left( \widehat{x} \right) \right] \nonumber \\
     & + \text{log} \left[D_{I} \left(x_p \right) \right]. 
 \end{align}
\normalsize

\noindent\textbf{Motion Learning.}
We train the Motion Stream to generate the continuous motion for the next 10 frames after observing the first frame. During training, the Motion Stream reconstructs the whole sequence recurrently. For every time step $t$, it reconstructs the current frame $\widehat{x}_{t}$ from $ \left \{x_1, x_2, \cdots, x_{t} \right \}$. 

Similarly to content learning, we optimize the motion stream with a motion VAE loss and a video-level GAN loss. In addition to the video reconstruction loss and KL-divergence between the prior motion latent distribution $ p\left(\mathbf{z_m} |k\right)$ and the posterior distribution $q \left(\mathbf{z_m} | \Delta x_t, k \right)$ where $\Delta x_t = x_t -x_{t-1}$, the VAE loss also contains a $\mathcal{L}_2$ loss between the refined content hidden layer of the previous frame $\left(\widehat{h_{c}^s}\right)_{t-1}$ and the content layer of the current frame $\left(h_{c}^s\right)_{t}$. Thus, the total motion VAE loss is:
\small
\begin{align}
     \min\limits_{E_m, G_m} \mathcal{L}_{VAE-{m}} = & \ \lambda_3 \sum\limits_{s,t} \mathcal{L}_2 \left(\left(\widehat{h_{c}^s}\right)_{t-1}, \Big(h_{c}^s \Big)_{t}\right) + \lambda_4 \mathcal{L}_2 \left(\widehat{v}, v \right)   \nonumber \\
     & + \lambda_{5} \sum\limits_{t} KL \left(q \left(\mathbf{z_m} | \Delta x_{t}, k \right)|| p \left(\mathbf{z_m}|k \right) \right).
\end{align}
\normalsize
The video-level GAN loss contains the vanilla GAN loss $\mathcal{L}_{GAN-{m}}$ and an auxiliary classification loss $\mathcal{L}_{cls}$ ~\cite{odena2016conditional}. The vanilla GAN takes  real videos $v$ as positive examples, and treats the generated videos $v_p$ and $\widehat{v}$ sampled from $p \left(\mathbf{v_c} | k \right)$ and $q \left(\mathbf{v_c} | x,k \right)$ as negative examples. Apart from GAN training, both the generator and the discriminator minimize a classification loss on action labels to generate  action-related motion:
\small
\begin{align}
     \max\limits_{G_m} \min\limits_{D_V} \mathcal{L}_{GAN-m} = & \  \text{log} \left[1 - D_{V} \left(v \right) \right] +  \text{log}\left[ D_{V}\left(\widehat{v}\right)\right] \nonumber \\
       & +  \text{log} \left[ D_{V} \left(v_p \right ) \right] , \\
     \min\limits_{G_m, D_V} \mathcal{L}_{cls} \left(v\right) + \mathcal{L}_{cls} & \left(\widehat{v} \right)  + \mathcal{L}_{cls} \left (v_p \right). 
\end{align}
\normalsize

We provide implementation details (hyper-parameters, model architecture, etc.) in the supplementary material.

\subsection{Generating a Video at Test Time}
While learning relies on observing the ground truth, at test time, generating a video begins from sampling in the latent space.
Given an action class $k$, we generate the first frame from a randomly sampled content vector $\epsilon_c \sim p\left(\mathbf{z_c} |k\right)$, and then  the following frames from the content embedding of the last frame as well as the current motion embedding $e_m$ computed by the convLSTM recurrently. To generate $e_m$ at each time step, the convLSTM updates the accumulated motion history with an extra motion vector $\epsilon_m \sim p\left(\mathbf{z_m} | k\right)$ containing the current potential motion.


\section{Experiments}
\subsection{SynAction Dataset} \label{sec:syn_action} 
        

\begin{table}[]
    \centering
    \resizebox{\linewidth}{!}{
    \begin{tabular}{c| c c c c c }
        
        \hline
         Dataset & \textbf{SynAction} & Moving MNIST & Shape Motion & Weizmann & MUG \\
        \hline
        Type & Synthetic & Synthetic & Synthetic & Real & Real \\
         Videos & 6000 & 10000 & 4000 & 90 & 882 \\
         Action Models & 120 & - & 2 & - & -\\
         Action Classes & 20 & 1 & 1 & 10 & 6 \\ 
         Scenes & 150 & 10 & 1 & 9 & 52 \\
         \hline
    \end{tabular}}
   \caption{ \small Compared to other datasets frequently used in video generation, our SynAction Dataset 
   has a larger variation of content and motion and is thus more challenging.}
    \label{tab:synaction}
    \vspace{-15pt}
\end{table}

Existing synthetic datasets used in \cite{srivastava2015unsupervised, xue2016visual, tulyakov2017mocogan} only contain rigid motions (e.g. linear motion in any direction) which are not challenging for deep neural networks. To uncover the model's ability to generate complex and action-related motion, we build a large-scale synthetic human action dataset, \textit{SynAction Dataset}, with a powerful game engine \textit{Unity}.

The dataset contains 120 unique non-rigid human action models from the Mixamo motion library across 20 action classes. Every action is akin to real human actions but easy to distinguish from other actions. The dataset is further varied between 10 different actors and 5 different backgrounds.
Table~\ref{tab:synaction} shows that SynAction has more variation in content and motion than existing synthetic and standard real-world datasets for video generation.

We provide each video with four different annotations: actor identity, action class, background and viewpoint. In this paper, we only use the action class to generate videos. 

\begin{table*}
    \begin{center}
        \resizebox{0.7\linewidth}{!}{
        \begin{tabular}{|c|ccc|ccc|ccc|}
             \hline 
             Dataset & \multicolumn{3}{c|}{Weizmann (\# action = 10)} &\multicolumn{3}{c|}{MUG (\# action = 6)} & \multicolumn{3}{c|}{Syn-Action (\# action = 20)} \\
             \hline
             Metric   & $\text{H} \left(y \right)$   & $\text{H} \left(y|v \right)$ & IS  & $\text{H} \left(y \right)$  & $\text{H}\left(y|v \right)$  & IS & $\text{H} \left (y \right)$ & $\text{H} \left (y|v \right )$ & IS  \\
             \hline
             MoCoGAN \cite{tulyakov2017mocogan} 
             & 4.38 & 0.31 &  58.52 & 
             1.78 & 0.17  & 5.03 &
             2.90 & 0.36 & 12.75 \\
             VideoVAE \cite{he2018probabilistic} 
             &  4.37 & 0.11 & 70.10 &
             - & - & -  & 
             - & - & -  \\
             SGVAN 
             & 4.34 & \textbf{0.04} & 73.73 &
             1.79 & 0.13 & 5.29 & 
             2.98 & 0.18 & 16.43 \\
             TwoStreamVAN ($-$M) 
             & 4.31 & 0.29 & 55.99 & 
             1.79 & 0.11 & 5.32  & 
             2.97 & 0.15 & 16.79 \\
            TwoStreamVAN 
             & \textbf{4.40} & 0.05 & \textbf{77.11} &
             \textbf{1.79} & \textbf{0.09} & \textbf{5.48}  & 
             \textbf{2.99} & \textbf{0.09} & \textbf{18.27} \\
            \hline 
            \textit{Exp Bound}
             &  4.50 & 0.01 & 88.94 &
             1.79 & 0.01 & 5.91 & 
              3.00 & 0.01 & 19.85 \\
             \textit{Math Bound}
              & 4.50 & 0.00 & 90.00 &
              1.79 & 0.00 & 6.00 & 
              3.00 & 0.00 & 20.00 \\
            \hline 
        \end{tabular}}
        \caption{\small Quantitative Results on Weizmann, MUG and Syn-Action Datasets. For IS, the higher value is better; while for $\text{H} \left(y|v \right)$, the lower value is better. Compared with all baselines, our TwoStreamVAN model achieves the best results on most metrics.}
        \label{table:quant_result}
    \end{center}
 
 \vspace{-20pt}
\end{table*}

\subsection{Other Datasets} \label{sec:standard_datasets}
In addition to our proposed SynAction Dataset, we evaluate our model on three existing datasets: Weizmann Human Action~\cite{ActionsAsSpaceTimeShapes_pami07}, MUG Facial Expression~\cite{aifanti2010mug} and VoxCeleb~\cite{Nagrani17}.
The Weizmann Human Action Dataset contains 90 videos of 9 actors performing 10 different actions. MUG Facial Expression Dataset contains 882 videos with 52 actors performing 6 different facial expressions. From the VoxCeleb Dataset, we form a training set containing 15184 videos of 186 people speaking.

With these three datasets, we cover a large range of motion, from large human actions (e.g. running, jumping) to subtle facial movements (e.g. happiness, disgust, speaking) as well as head movements (e.g. nodding, turning) and include both periodic and non-periodic motion.  

\subsection{Evaluation Metrics} \label{sec:evaluation_metrics}

Quantitative evaluation of generative models remains a challenging problem, and there is no consensus on the measurement which best evaluates the realism and diversity of the generated results. Thus, instead of just relying on a single measurement, we utilize four different metrics to examine both the realism and diversity of generated motion: Inception Score (IS) \cite{salimans2016improved}, Inter-Entropy $\text{H} \left(y\right)$  \cite{he2018probabilistic} and Intra-Entropy \cite{he2018probabilistic} $\text{H} \left (y|v \right)$, where $v$ is the video for evaluation and $y$ is the action predicted by a classifier. Because all these metrics utilize a pre-trained classifier for evaluation,
we train a classifier separately on each dataset and show its performance by computing the same metrics on the test set, which only consists of real videos.
We call these values the \textit{Experimental Bound}.
To make a fair comparison, we compute metrics on 10-frame video clips generated by each model.

\subsection{Baselines} \label{sec:baselines}
We compare against two existing methods to show our model's superiority in generating videos of a given action:
MoCoGAN\footnote{We use categorical MoCoGAN implemented by its authors.}~\cite{tulyakov2017mocogan} and VideoVAE\footnote{Due to VideoVAE's non-public implementation, we only compare with quantitative results on Weizmann Dataset reported in the paper.}~\cite{he2018probabilistic}, which are the current state-of-the-art. 

We also design several ablated variants of TwoStreamVAN to examine key components of our model:

\textbf{SGVAN} adopts a single generator to generate a single frame from the disentangled content and motion vectors, keeping all parts of the model the same. This comparison evaluates the contribution of the parallel $G_c$ and $G_m$.

\textbf{TwoStreamVAN($-$M)} applies motion fusion to content hidden layers at multiple scales without the guidance of motion masks.
This comparison helps us to examine the effectiveness of motion masks.

\subsection{Results} \label{sec:quant_results}
\vspace{-2mm}
\paragraph{Quantitative Results.}
We compute quantitative metrics on the results of all baselines and our TwoStreamVAN (see Table.~\ref{table:quant_result}) on Weizmann, MUG and SynAction Datasets. We train a normal action classifier on MUG and SynAction Datasets, and train a classifier to distinguish each actor-action pair on Weizmann to compare with VideoVAE. 

TwoStreamVAN improves the Inception Scores of MoCoGAN by $32\%$, $9\%$ and $43\%$ on Weizmann, MUG and SynAction Datasets respectively. Despite that VideoVAE receives the first frame and the actor identity as additional inputs, our model still results in a better IS value on Weizmann. Meanwhile, TwoStreamVAN achieves both higher $\text{H}(y)$ and lower $\text{H}(y|v)$, indicating that it generates more diverse and more realistic videos than either MoCoGAN or VideoVAE. 

Compared to our ablated models SGVAN and TwoStreamVAN($-$M), our model pushes all metrics closer to their bounds. These results reveal that our full model benefits from key components in our design, namely the disentangled generators and the guidance of the motion mask in fusion.

\begin{figure*}
    \centering
    \includegraphics[width=0.9\linewidth]{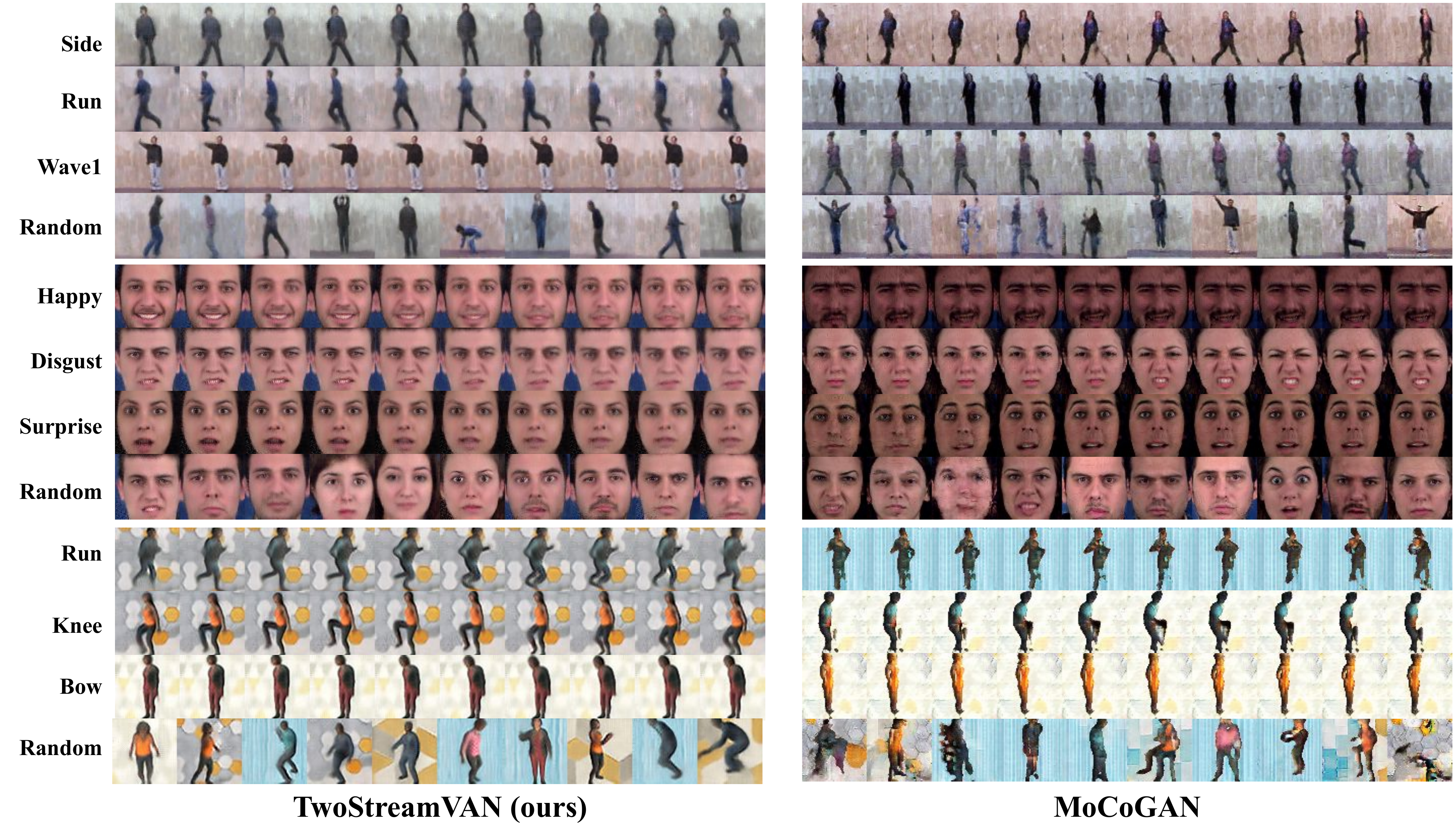}
    \vspace{-5pt}
    \caption{\small We provide 3 generated videos of TwoStreamVAN and MoCoGAN on Weizmann, MUG and our new SynAction datasets. Our model generates clear and consistent motion. We further randomly sample 10 generated frames for each model. TwoStreamVAN results in better content quality with fewer distortions in every single frame. More qualitative results are presented in the supplementary material.}
    \label{fig:qual_seq}
    \vspace{-15pt}
\end{figure*}

\vspace{-12pt}
\paragraph{Qualitative Results.} \label{sec:qual_results}
We visualize videos generated by TwoStreamVAN and MoCoGAN. For Weizmann, MUG and SynAction datasets, we provide 4 generated videos from TwoStreamVAN and MoCoGAN of the given action
(Fig.~\ref{fig:qual_seq}). Our TwoStreamVAN model succeeds in generating more accurate and fine-grained motion for different actions. 

To evaluate the quality of content generation, we randomly sample 10 generated frames from TwoStreamVAN and MoCoGAN's results respectively (Fig.~\ref{fig:qual_seq}). In comparison, TwoStreamVAN yields better content generation, with few severe distortions or bizarre artifacts across all three datasets, which demonstrates that TwoStreamVAN reduces the content deterioration significantly.

\vspace{-12pt}

\paragraph{User Study on SynAction.} \label{sec:user_study} 
To further test the ability of TwoStreamVAN and MoCoGAN to handle more diverse human action videos, we conduct user studies on SynAction via AMTurk~\cite{buhrmester2011amazon}.
In the pairwise comparison, $\mathbf{88\% \pm 0.45 \%}$ / $\mathbf{81\% \pm 0.71 \%}$ users think TwoStream generates better videos before/after knowing the ground truth action, showing that the superiority of our TwoStreamVAN to generate the clear and consistent motion and yield visually satisfying videos.

\begin{figure}
    \centering
    \includegraphics[width=0.9\linewidth]{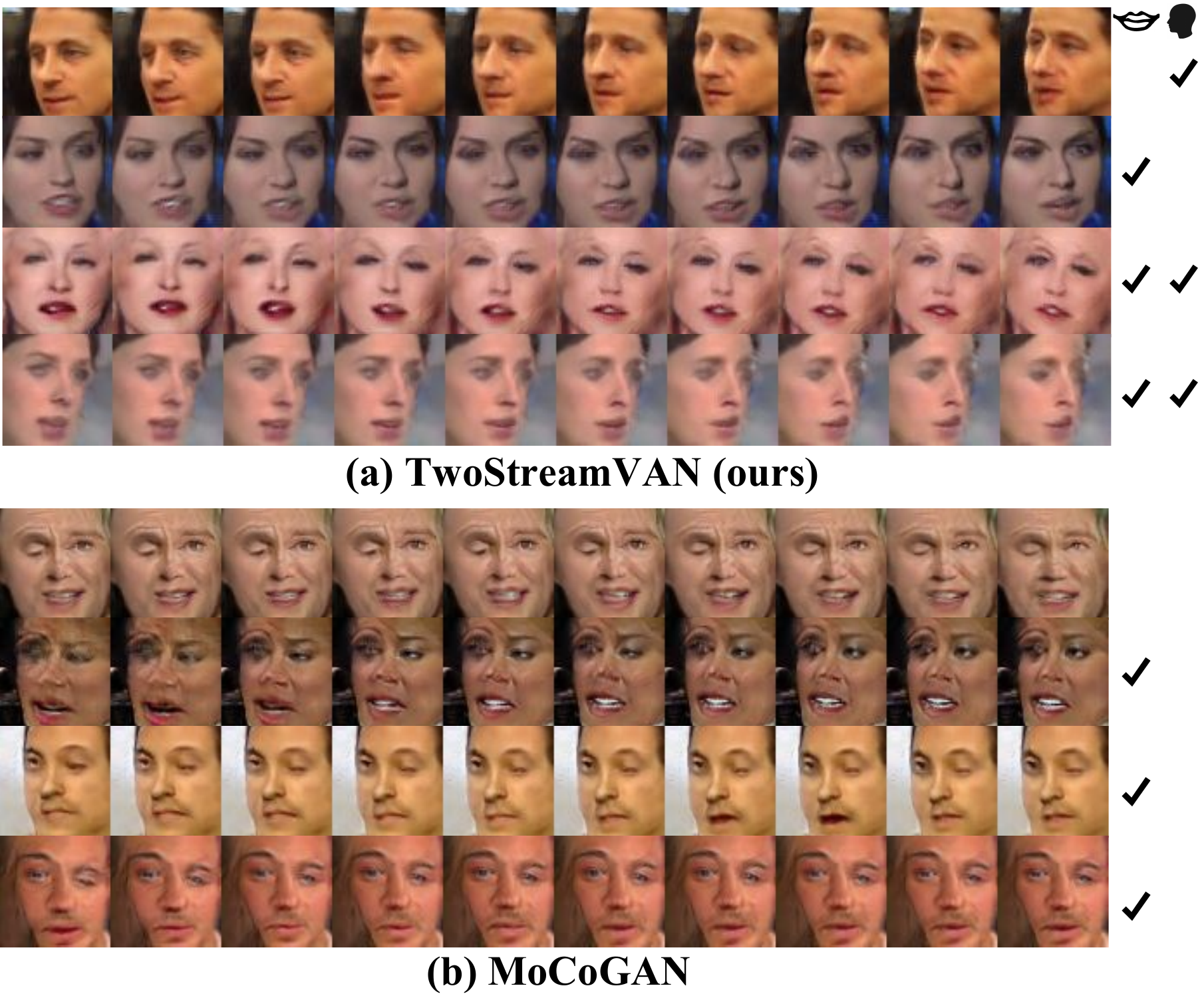}
    \vspace{-10pt}
    \caption{\small Qualitative comparison between TwoStreamVAN and MoCoGAN. TwoStreamVAN captures both lip and head movements, while MoCoGAN only generates non-moving heads. More examples are provided in the supplementary material.}
    \label{fig:voxceleb_qual}
    \vspace{-20pt}
\end{figure}

\vspace{-12pt}
\paragraph{Experiment on large-scale VoxCeleb.} 
To extensively test the ability of our model to generate real-world videos, we experiment on a new large-scale speech dataset VoxCeleb~\cite{Nagrani17}. 
Compared to the standard Weizmann and MUG Datasets, VoxCeleb is more challenging due to the wide variety of content and motion in its huge amount of videos.
Because VoxCeleb has no action label, we generate videos only from a noise vector. We compare to MoCoGAN via a user study on AMTurk and the qualitative visualization. 

In the pairwise comparison, $\mathbf{82.55\% \pm 0.53\%}$ users prefer TwoStreamVAN to MoCoGAN, evidence that our generated videos are more visually-pleasing. In the qualitative comparison (see Fig~\ref{fig:voxceleb_qual}), TwoStreamVAN captures both lip and head movements, whereas MoCoGAN suffers from mode collapse and only generates videos without head movements. Our model precisely reproduces the subtle motion related to speaking, while MoCoGAN outputs exaggerated and unrealistic movements (e.g. teeth suddenly appear during speech). Moreover, TwoStreamVAN generates better-looking faces than MoCoGAN.

\subsection{Ablation Studies}
\begin{figure}
	\centering
   \includegraphics[width=\linewidth]{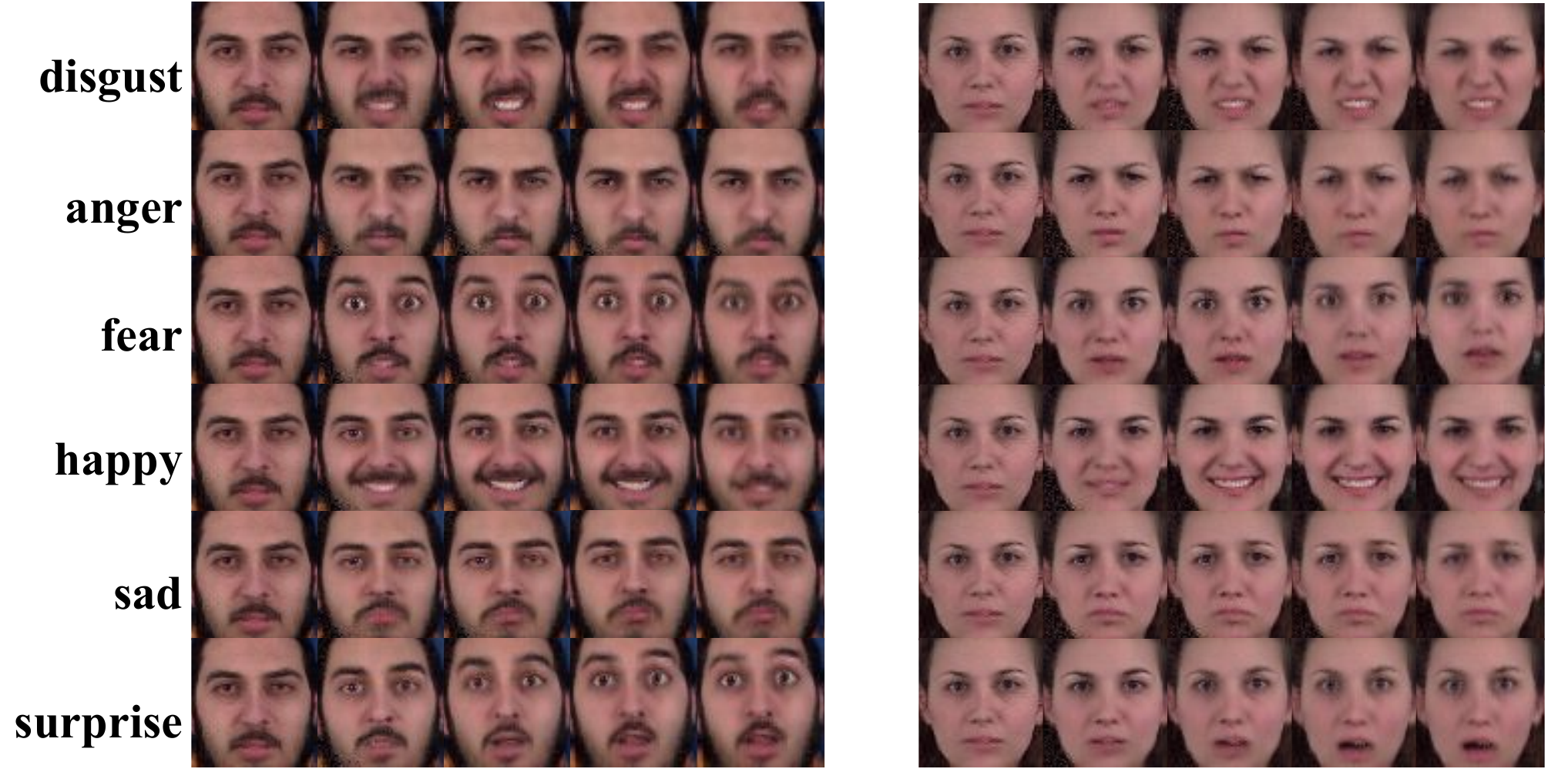}
   \vspace{-15pt}
	\caption{\small We provide examples of varying the motion code while the content code is fixed.}
    \label{fig:disentanglement}
  \vspace{-15pt}
\end{figure}

\paragraph{Disentanglement Study}
To show that the disentanglement of content and motion really happens, we experiment on MUG dataset. With the fixed content code $\epsilon_c$, we generate videos using different motion codes (see Fig~\ref{fig:disentanglement}). We classify actors and actions in the generated videos using pre-trained classifiers.  $95.9\%$ of our generated videos preserve the actor in the fixed content code and $95.6\%$ of videos generate the correct action defined in the different motion code. It demonstrates that content and motion generations are separated in TwoStreamVAN.

\paragraph{Multi-scale v.s. Single-scale Motion Fusion} \label{sec:multi-scale_ablation}
    
To examine the effectiveness of multi-scale motion fusion, we train four TwoStreamVAN models on the Weizmann Human Action Dataset, in which we apply fusions at 1, 2, 3 and 4 layers (scales) respectively. We add new fusion layers from the highest resolution to the lowest resolution. In each fusion, we implement motion kernels with a fixed size $n=5$.  Moreover, we train a model where we apply large motion kernels with $n=17$ on the output image from the Content Stream to imitate the single-scale fusion in \cite{niklaus2017video}.

In Table.~\ref{tab:fusion_quant}, it is not surprising that the performance drops when we reduce $n$ from 17 to 5 with a single fusion on the image. As we increase fusion layers, IS value, $\text{H} \left(y \right)$ and $\text{H} \left(y|v \right)$ recover and finally outperform those of the model with a single large fusion step on the full resolution image. 

\begin{figure}
    \centering
    \includegraphics[width=\linewidth]{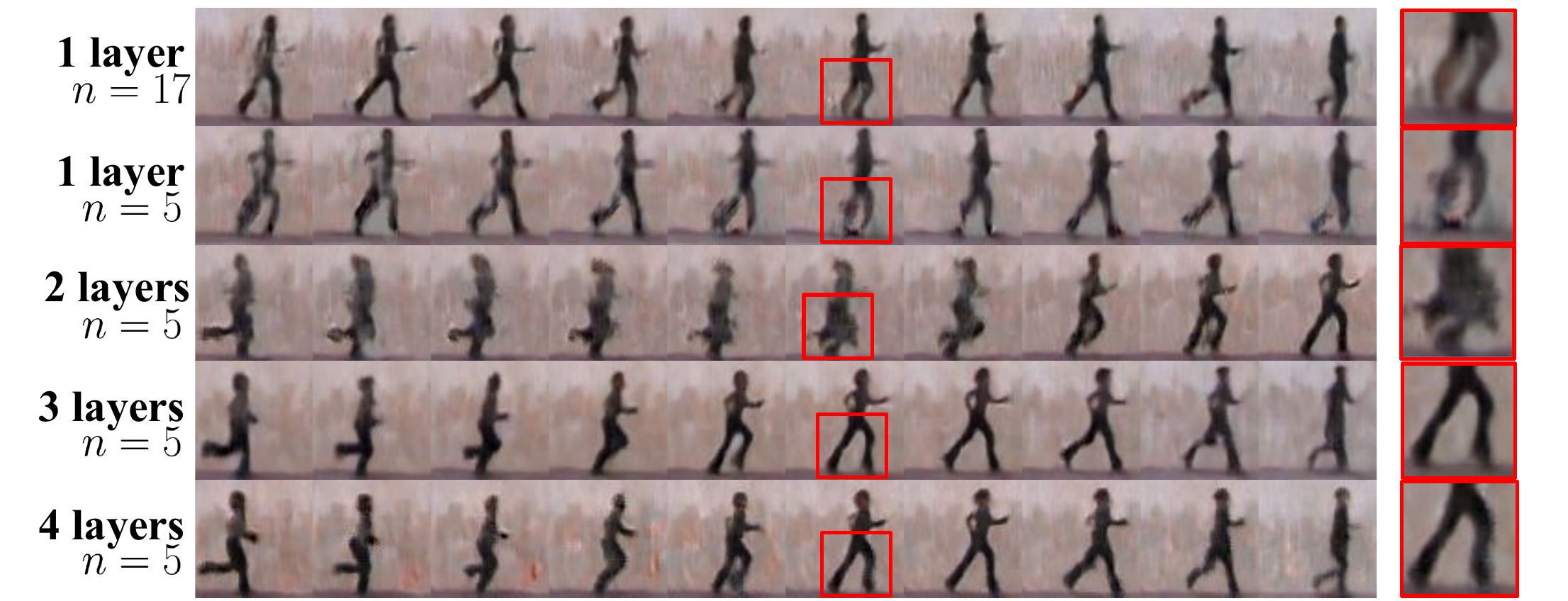}
    \vspace{-15pt}
    \caption{\small Qualitative comparison among similar sequences generated by single fusion with large kernels and multi-scale motion fusion with small kernels. When we apply fusion at 3 or 4 scales, our model generates the sharpest and clearest outlines of the runner's leg among all models.}
    \label{fig:fusion_qual}
    \vspace{-10pt}
\end{figure}

To further analyze multi-scale fusion, we measure $\text{H} \left(y|v \right)$ of large motion, e.g. running (in Table.~\ref{tab:fusion_quant}). The more layers such fusion is applied to, the lower $\text{H} \left (y|v \right)$ is, indicating that more realistic large motion is generated. We pick similar videos (Fig.~\ref{fig:fusion_qual}) generated by different models and zoom in on the actor's legs, where the largest motion happens. When we only apply fusion with kernel size $n=5$ at the highest 2 resolutions, the model fails to tackle the large motion around the legs and generates blobs. After increasing the number of fusion layers, it finally generates an even sharper outline than the model using the single large fusion, showing the benefits of our multi-scale mechanism. 
    
\begin{table}[]
    \centering
    \resizebox{\linewidth}{!}{
    \begin{tabular}{|c|c|cccc|}
        \hline
        $\#$ layers & n & IS $\uparrow$ & $\text{H} \left(y \right)$ $\uparrow$ & $\text{H} \left(y|v \right)$ $\downarrow$ & $\text{H} \left(y|v \right)$ (run) $\downarrow$ \\
        \hline
        1 & 17 & 75.89 & 4.38 & \textbf{0.05} & 0.104 \\
        \hline
        1 & 5 & 74.95 & 4.39 & 0.07 & 0.166 \\
        2 & 5 & 74.57 & 4.38 & 0.07 & 0.096 \\
        3 & 5 & \textbf{77.83} & \textbf{4.40} & \textbf{0.05} & 0.062 \\
        4 & 5 & 77.11 & \textbf{4.40} & \textbf{0.05} & \textbf{0.059} \\ 
        \hline
    \end{tabular}}
    \caption{ \small Quantitative results of models varying in the kernel size $n$ and the number of scales to apply fusion. For single fusion, the performance drops by reducing the kernel size. After applying fusion at multiple scales, the performance recovers and finally out-performs the single fusion with large kernels.}
    \label{tab:fusion_quant}
    \vspace{-10pt}
\end{table}

\begin{figure}
	\centering
   \includegraphics[width=\linewidth]{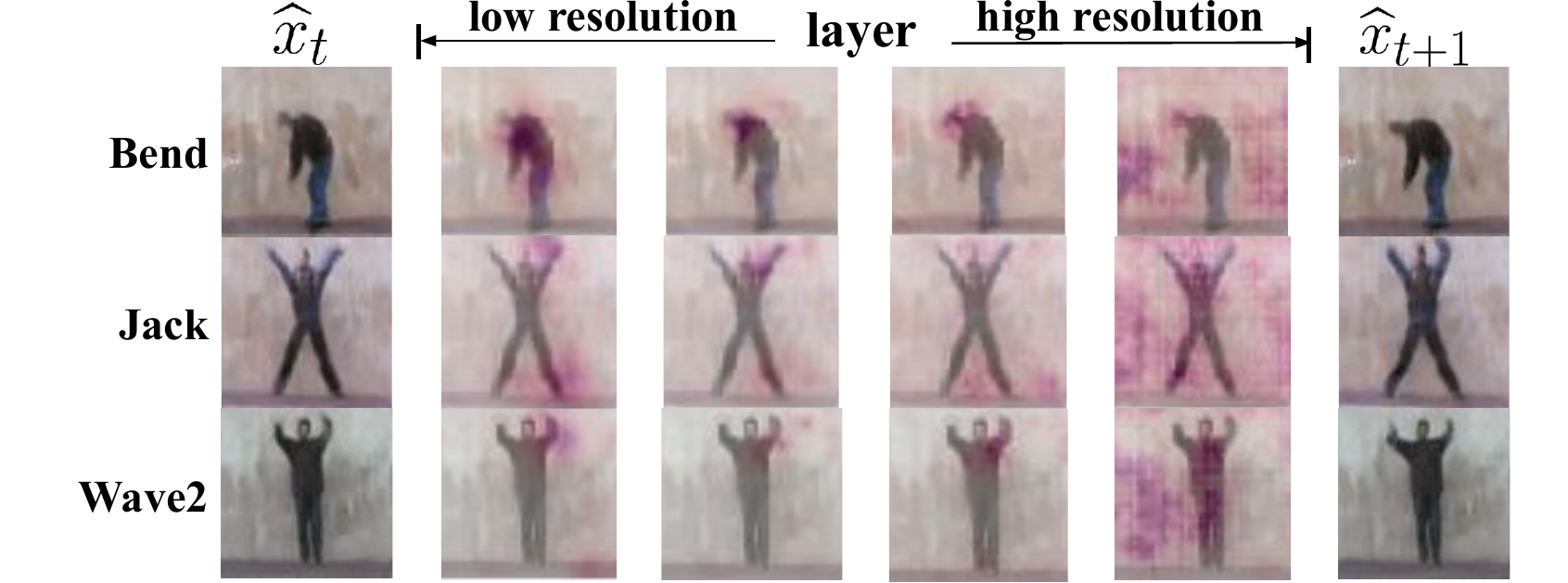}
   \vspace{-15pt}
    \label{fig:mask_overlay}
	\caption{\small We overlay the motion mask at each scale with the current frame $\widehat{x}_t$. Motion masks at lower resolutions are only activated on large-motion areas, while a large area in the motion mask at the highest resolution is activated to tackle small changes between neighboring frames.}
    \label{fig:mask_overlay}
  \vspace{-15pt}
\end{figure}

\begin{figure}[t]
	\centering
    \includegraphics[width=\linewidth]{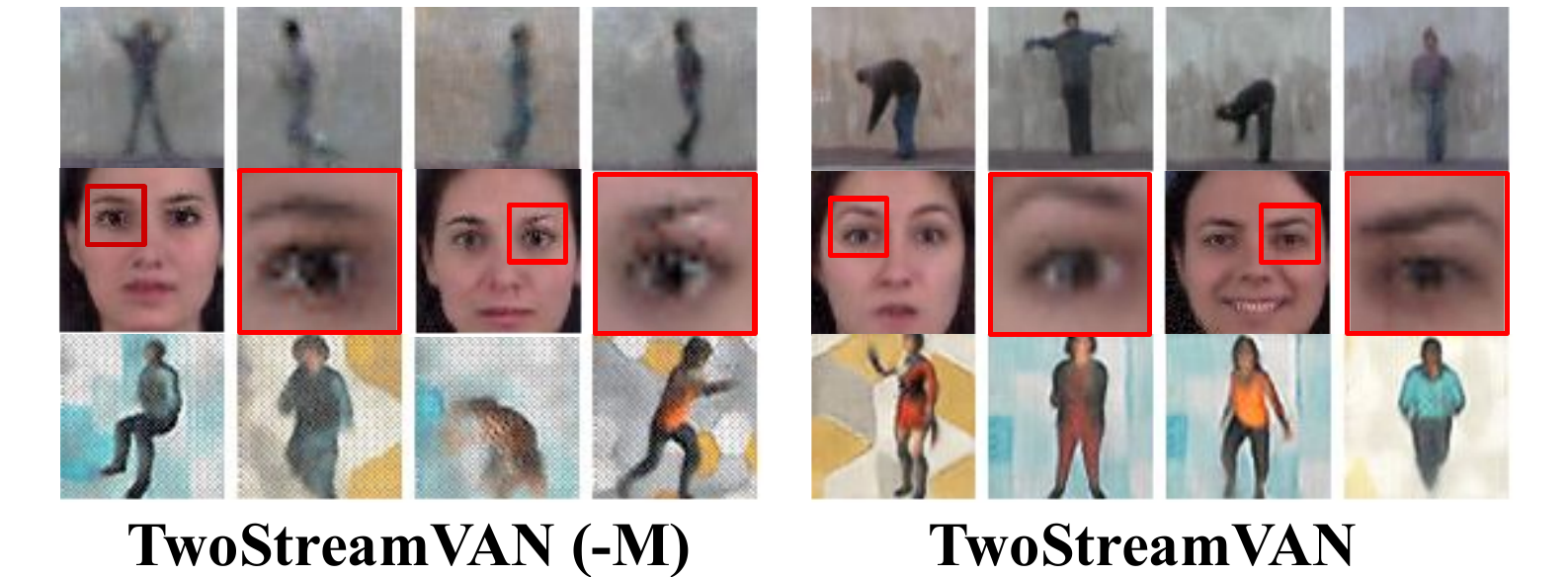}
    \vspace{-15pt}
    \caption{\small Qualitative Comparison of TwoStreamVAN($-$M) and TwoStreamVAN. Motion masks improve generation in a few areas (background for Weizmann and SynAction and eyes for MUG).}
    \label{fig:qual_nomask}
    \vspace{-15pt}
\end{figure}

\vspace{-10pt}
\paragraph{Visualization of Motion Masks} \label{sec:mask_visual}
We visualize the motion mask at each layer by overlaying it with the current frame $\widehat{x}_t$ (Fig.~\ref{fig:mask_overlay}) to show it activates the correct area. For low resolutions, the mask is only activated at large-motion areas, e.g. torso for bending, arms and legs for jumping-jacks and arms for waving. For the highest resolution, the activation covers the background area to overcome some small changes, e.g. lighting changes or small camera movements. This correct activation focuses the model's attention on learning the motion in these areas and boosts quantitative performance (see TwoStreamVAN~($-$M) v.s. TwoStreamVAN in Sec.~\ref{sec:quant_results}). 

We further visualize frames generated from these two models (Fig.~\ref{fig:qual_nomask}). We observe that TwoStreamVAN($-$M) does a worse job in small-motion areas. On Weizmann and SynAction, it messes up background patterns. On MUG, it generates unexpected brick patterns around eyes. Our full TwoStreamVAN does not suffer from these problems with the help of motion masks. This observation is consistent with our claim (in Sec.~\ref{sec:motion_refinement}) that motion masks help to preserve static pixel values during the generation.

\section{Conclusion}
In this paper, we propose a novel Two-Stream Variational Adversarial Network to improve motion modeling in video generation. To generate motion efficiently, we decompose content and motion in the generation phase and fuse them via a novel multi-scale mechanism. Combined with the dual-task learning scheme, our disentangled generative network overcomes the common content deterioration and further benefits motion modeling. We propose a large-scale synthetic human action dataset \textit{SynAction} to evaluate the motion modeling in video generation. Our model significantly outperforms the current state-of-the-art works across Weizmann, MUG, VoxCeleb and our SynAction datasets in quantitative and qualitative (user study) evaluations. In the future work, we hope to explore video generation controlled via textual description or interactive user manipulation.

\section{Acknowledgements}
This work is supported by DARPA LWLL Project. It reflects the opinions and conclusions of its authors, but not the funding agents.

\newpage

\appendix
\noindent Please also refer to \href{https://youtu.be/LNKwA9KsoJ0}{our video comparison} for more details.
\section{Implementation Details}

We implement our model using PyTorch~\cite{paszke2017automatic}. We use Xavier initialization~\cite{glorot2010understanding} for each layer and use the Adam optimizer~\cite{kingma2014adam} with initial learning rate $\alpha = 2 
\times 10 ^{-4}$, first decay rate $\beta_1 = 0.5$ and second decay rate $\beta_2=0.999$. We train our model for a total of 500K iterations with batch size 16 which takes 2 days on a TITAN V GPU to finish, and the ratio between Content and Motion Learnings is 3:2. To generate more complicated content in Syn-Action and Datasets, we pre-train the Content Stream for 300K iterations by the image reconstruction task. At the test time, we heat up the network for two time steps before generating videos on each dataset.

Inspired by the curriculum learning approach~\cite{bengio2015scheduled} and the scheduled sampling mechanism~\cite{bengio2015scheduled}, we design the motion learning as follows. We introduce a very simple learning task at a very early stage, where the Motion Stream is trained to predict the next frame sorely from the current frame with no need of modeling the history. This task is gradually replaced by the sequence training task using the scheduled sampling strategy such that at the beginning the model is trained for one-step prediction providing the entire history, while by the end of training the model is fully auto-regressive. 

\subsection{Model Architecture}
\begin{figure}
    \centering
    \includegraphics[width=0.9 \linewidth]{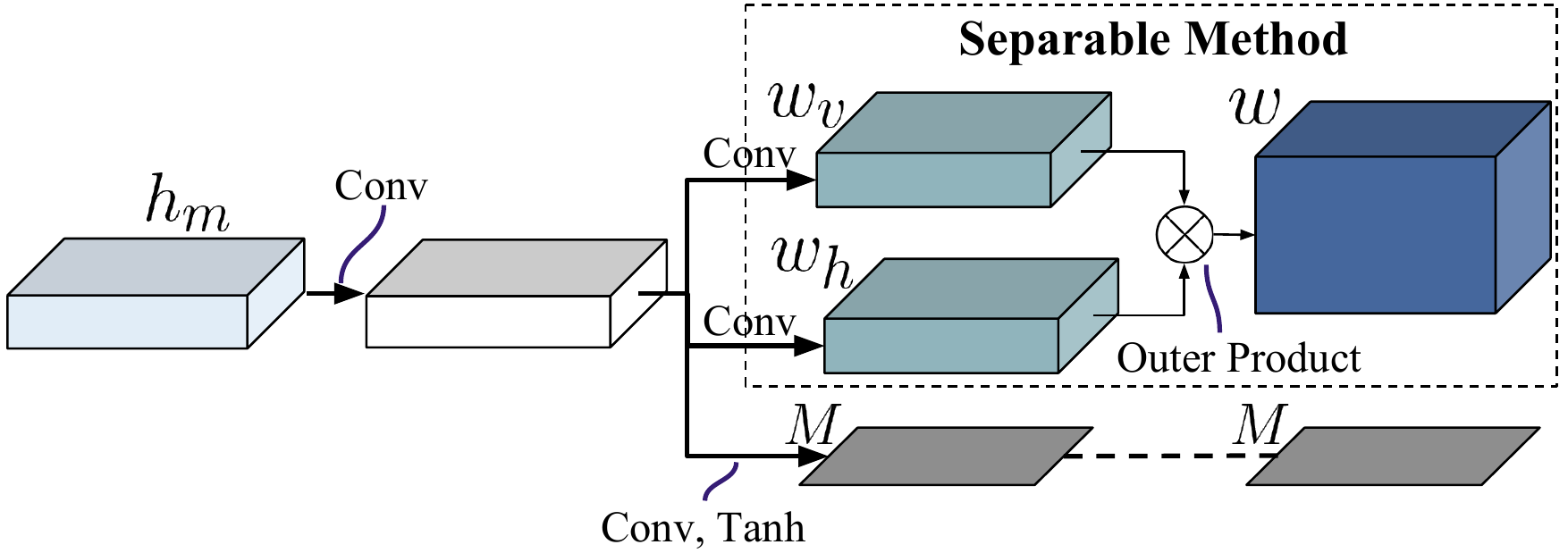}
    \caption{We design a subnet to generate motion kernels $w$ and the motion mask $M$ at each scale. $h_m$ is a fraction of motion feature map which is used to generate $w$ and $M$. In the separable method, a 2D kernel $w(a, b)$ is approximated by the outer-product of two 1D kernels: $w_v(a,b)$ and $w_h(a,b)$. This subnet contains three convolutional branches to generate $w_v$, $w_h$ and $M$ respectively.}
    \label{fig:subnet}
\end{figure}

We provide the detailed model architecture of the Content Stream (the Content Encoder $E_c$, the Content Generator $G_c$ and the Image Discriminator $D_I$) in Table.~\ref{tab:model_arch_content} and the Motion Stream (the Motion Encoder $E_m$, the Motion Generator $G_m$, the Video Discriminator $D_V$) in Table.~\ref{tab:model_arch_motion}. In $E_c$ and $G_c$, we do not apply the Batch Normalization~\cite{ioffe2015batch} after each convolutional layer to stabilize the content generation. For each scale $s$ in the Motion Stream, we design a small subnet (Fig.~\ref{fig:subnet}) to generate adaptive convolution kernels $w^s$ and $M^s$ from a fraction of the motion feature map $h_m$. The weights of the subnet are not shared by different scales.  Additionally, we generate $w^s$ from two separate 1D kernels $w_h^s$ and $w_v^s$ as in \cite{niklaus2017videoseparable}. Our implementation will be available.

\begin{table}[]
    \centering
    \begin{tabular}{c|l}
        \noalign{\hrule height 1pt}
        layer & configuration \\
        \noalign{\hrule height 1pt}
        \multicolumn{2}{c}{Content Encoder $E_c$} \\
        \noalign{\hrule height 1pt}
        1 & Conv2D (\ngf, 3, 2, 1), ReLU \\
        \hline
        2 & Conv2D (\ngf, 3, 1, 1), ReLU \\
        \hline
        3 & Conv2D (2x\ngf, 3, 2, 1), ReLU \\
        \hline
        4 & Conv2D (2x\ngf, 3, 1, 1), ReLU \\
        \hline
        5 & Conv2D (\ngf, 3, 2, 1), ReLU \\
        \hline
        6 & FC (32x\ngf), ReLU, FC(2x$\mathcal{C}$) \\
        \noalign{\hrule height 1pt}     
        \multicolumn{2}{c}{Content Generator $G_c$} \\
        \noalign{\hrule height 1pt}
        1 & FC(32x\ngf),ReLU, FC(64x\ngf), ReLU \\
        \hline
        2 & Deconv2D (8x\ngf, 3, 2, 1), ReLU \\
        \hline
        3 & Deconv2D (4x\ngf, 3, 1, 1), ReLU \\
        \hline
        4 & Deconv2D (2x\ngf, 3, 2, 1), ReLU \\
        \hline
        5 & Deconv2D (2x\ngf, 3, 1, 1), ReLU \\
        \hline
        6 & Deconv2D (\ngf, 3, 2, 1), ReLU \\
        \hline
        7 & Deconv2D (3, 3, 2, 1), Tanh \\
        \noalign{\hrule height 1pt}
        \multicolumn{2}{c}{Image Discriminator $D_I$} \\
        \noalign{\hrule height 1pt}
        1 & Conv2D (16, 4, 2, 1), LeakyReLU \\
        \hline 
        2 & Conv2D (32, 4, 2, 1), BN, LeakyReLU \\
        \hline 
        3 & Conv2D (64, 4, 2, 1), BN, LeakyReLU \\
        \hline
        4 & Conv2D (1, 4, 2, 1), Sigmoid \\
        \noalign{\hrule height 1pt}
    \end{tabular}
      \caption{Model Architecture of the Content Stream. For each convolution layer, we list the output dimension, kernel size, stride, and padding. For the fully-connected layer, we provide the output dimension. `\ngf' is the basic output dimension, which is a hyper-parameter in the model architecture. $\mathcal{C}$ is the dimension of the content latent space.}
    \label{tab:model_arch_content}
\end{table}

\begin{table}[]
    \centering
    \resizebox{\linewidth}{!}{
    \begin{tabular}{c|l}
        \noalign{\hrule height 1pt}
        layer & configuration \\
        \noalign{\hrule height 1pt}
        \multicolumn{2}{c}{Motion Encoder} \\
        \noalign{\hrule height 1pt}
        1 & Conv2D (\ngf, 5, 1, 2), BN, ReLU \\
        \hline
        2 & Conv2D (\ngf, 5, 1, 2), BN, ReLU, MaxPool \\
        \hline
        3 & Conv2D (2x\ngf, 5, 1, 2), BN, ReLU \\
        \hline
        4 & Conv2D (2x\ngf, 5, 1, 2), BN, ReLU, MaxPool \\
        \hline
        5 & Conv2D (2x\ngf, 7, 1, 3), BN, ReLU \\
        \hline
        6 & Conv2D (\ngf, 7, 1, 3), BN, ReLU, MaxPool \\
        \hline
        7 & FC (512), ReLU, FC(2x$\mathcal{M}$) \\
        \noalign{\hrule height 1pt}
        \multicolumn{2}{c}{Motion Generator} \\
        \noalign{\hrule height 1pt}
        0 & FC(1024), ReLU \\
        \hline
        1 & Deconv2D (128, 3, 1, 1), BN, ReLU \\
        \hline
        2 & Conv2D (128, 3, 1, 1), BN, ReLU \\
        \hline
        3 & Deconv2D (64, 4, 2, 1), BN, ReLU \\
        \hline
        4 & Conv2D (64, 3, 1, 1), BN, ReLU \\
        \hline
        5 & Deconv2D (32, 4, 2, 1), BN, ReLU \\
        \hline
        6 & Conv2D (32, 3, 1, 1), BN, ReLU \\
        \hline
        7 & Deconv2D (16, 4, 2, 1), BN, ReLU \\
        \hline
        8 & Conv2D (16, 3, 1, 1), BN, ReLU \\
        \noalign{\hrule height 1pt}
        \multicolumn{2}{c}{Motion Generator-Subnet} \\
        \noalign{\hrule height 1pt}
        1 & Deconv2D (16, 3, 1, 1), BN, ReLU \\
        \hline
        2-1 & Conv2D (5, 3, 1, 1) \\
        \hline
        2-2 & Conv2D (5, 3, 1, 1) \\
        \hline
        2-3 & Conv2D (1, 3, 1, 1), Tanh \\
        \noalign{\hrule height 1pt}
        \multicolumn{2}{c}{Video Discriminator} \\
        \noalign{\hrule height 1pt}
        1 & Conv3D (64, 4, (1, 2, 2), (0, 1, 1)), LeakyReLU \\
        \hline
        2 & Conv3D (128, 4, (1, 2, 2), (0, 1, 1)), BN, LeakyReLU \\
        \hline
        3 & conv3D (256, 4, (1, 2, 2), (1, 1, 1)), BN, LeakyReLU \\
        \hline
        4 & Conv2D (512, 4, (1, 2, 2), (1, 1, 1)), BN, LeakyReLU \\
        \hline
        5 & Conv2D (512, 4, (1, 2, 2), (0, 1, 1)) \\
        \hline
        \noalign{\hrule height 1pt}
    \end{tabular}
    }
    \caption{Model Architecture of the Motion Stream. For each convolution layer, we list the output dimension, kernel size, strides, and padding. For each fully-connected layer, we provide the output dimension. `\ngf' is the basic output dimension, which is a hyper-parameter in the model architecture. $M$ is the dimension of the motion latent space. }
    \label{tab:model_arch_motion}
\end{table}

\subsection{Hyper-Parameter Setting}
We provide hyper-parameters of loss functions and model architecture for Weizmann Human Action~\cite{ActionsAsSpaceTimeShapes_pami07}, MUG Facial Expression~\cite{aifanti2010mug}, VoxCeleb~\cite{Nagrani17} and Syn-Action Datasets (Table.~\ref{table:hyper_parameters}). Since we adopt the scheduled sampling mechanism in the Motion Learning, we slowly increase $\lambda_6$ along with the process of the scheduled sampling, to restrict the KL divergence between the approximated latent distribution $q(\mathbf{z_m}|\Delta x, k)$ and the real latent distribution $p(\mathbf{z_m}|k)$ within a reasonable range. This helps to stabilize the motion sampling at the test time. We tune hyperparameters via validation: choose the best metrics computed from a group of generated videos. Note that the metrics reported in the paper are computed from another group of generated videos. We fix $\lambda_1$, $\lambda_3$ and $\lambda_4$, and sweep $\lambda_2$ in $[3, 8]$, `ngf` from $\{16, 32, 64\}$, $\mathcal{C}$ from $\{256, 512, 1024\}$ lower bound of $\lambda_5$ in $[1, 10]$ and higher bound from 5 to 10 times of its lower bound.

\begin{table}
    \begin{center}
        \resizebox{\linewidth}{!}{
        \begin{tabular}{|c|cc|ccc|ccc|}
             \hline 
              & \multicolumn{2}{c|}{Content Loss}
              & \multicolumn{3}{c|}{Motion Loss}
              &\multicolumn{3}{c|}{Model Arch} \\
             \hline
             Params &  $\lambda_1$  &  $\lambda_2$   &  $\lambda_3$ &  $\lambda_4$ & $\lambda_5$  & \ngf & $\mathcal{C}$ & $\mathcal{M}$ \\
             \hline
            Weizmann & $10^4$ & 7 &  $10^2$ & $10^4$ & $ 2\rightarrow20 $ & 16 & 512 & 64 \\  
            \hline
            MUG & $10^4$ & 5  & $10^2$ & $10^4$ & $ 5\rightarrow25 $ & 16 & 512 & 64 \\  
            \hline
            SynAction & $10^4$ & 7  & $10^2$ & $10^4$ & $ 2\rightarrow20 $  & 32 & 1024 & 64 \\  
            \hline 
            VoxCeleb & $10^4$ & 6 & $10^2$ & $10^4$ & $ 5\rightarrow25 $  & 64 & 512 & 64 \\  
            \hline 
        \end{tabular}
        }
     \caption{Hyper-Parameters for Weizmann Human Action, MUG Facial Expression, SynAction and VoxCeleb Datasets} \label{table:hyper_parameters}
    \end{center}
\end{table}


\section{Details of Experimental Setup}

\subsection{Data Spliting and Pre-processing}
We use three datasets to evaluate our model and other baselines: Weizmann Human Action~\cite{ActionsAsSpaceTimeShapes_pami07}, MUG Facial Expression~\cite{aifanti2010mug} and Syn-Action datasets.

\textbf{Weizmann Human Action.} Following  \cite{he2018probabilistic}, We use the first $2/3$ for the training and save the last $1/3$ for the test.

\textbf{MUG Facial Expression.} We use $4/5$ of the entire dataset for the training and save $1/5$ for the test.

\textbf{SynAction Dataset.} We use $14/15$ of the whole dataset for the training and save $1/15$ for the test.

\textbf{VoxCeleb Dataset.} We form the train set with $15184$ videos of $186$ people speaking. No test set is needed.

On all datasets, we crop the video centered at the actor or the face. To augment data, we further crop the video with a random small offset before down-sampling each frame to $64 \times 64$ at each iteration. We adjust the frame sampling rate based on action types to make motion observable between adjacent frames.

\subsection{Definition of Evaluation Metrics}
Let $v$ be the generated video and $y$ be the label for $v$, which is assigned by the pre-train classifier. We introduce definitions of Inter-Entropy $\text{H} \left(y \right)$, Intra-Entropy $\text{H} \left(y|v \right)$ and 
Inception Score (IS) and explain how they measure the diversity and realism of generative models. 

\textbf{Inter-Entropy} $\text{H} \left(y \right)$ is the entropy of the marginal distribution $p \left(y \right)$ obtained from all videos:
\begin{align}
    \text{H} \left(y \right) &= -\sum\limits_{y} p \left(y \right) \log p \left (y \right ), \\
    p \left (y \right ) & \approx \frac{1}{N} \sum\limits_{i=1}^{N} p \left(y|v_i \right).
\end{align}
If all classes are equally represented in the generated samples, $\text{H} \left(y \right)$ achieves its maximum value. Therefore, higher $\text{H} \left(y\right)$ indicates the model generates more diverse results.

\textbf{Intra-Entropy} $\text{H} \left(y|v \right)$ is the entropy of the conditional class distribution $p\left(y|v\right)$ of a single video $v$:
\begin{align}
    \text{H} \left(y|v \right) = -\sum\limits_{y} p \left(y|v \right)\log p \left (y|v \right),
\end{align}
 More confident the classifier is to predict its class, lower $\text{H} \left(y|v \right)$ is, and thus more realistic the video is. In this paper, we report the average $\text{H} \left(y|v \right)$ to evaluate the overall realism of the generated videos.

\textbf{Inception Score (IS)} is widely adopted to evaluate generative models. In video-level task, it measures the KL divergence between the conditional label distribution $p \left(y|v \right)$ and the marginal distribution $p\left(y \right)$:
\begin{align}
    \text{IS} & = \exp\left(\mathbb{E}_v \left[\mathrm{KL} \left(p \left(y|v \right) || p \left(y \right) \right) \right] \right) \nonumber \\
             & = \exp \left(\text{H} \left(y \right) - \mathbb{E}_v \left[\text{H} \left(y|v \right) \right] \right).
\end{align}
Inception Score favors a higher $\text{H} \left(y \right)$ and a lower $\text{H} \left(y|v \right)$. So it measures both the realism and diversity of the generated videos.

\subsection{Details of User Study}
To conduct the user study on SynAction and VoxCeleb datasets, we pairwisely compare 2000 random pairs via Amazon MTurk. Users are asked to choose a better looking one from a pair. For action-conditioned generation, users are further informed with the action label and re-make the choice. We calculate the percentage of user preference and bootstrap the variance.

\section{More Visualization Results}
\begin{figure*}
    \centering
    \includegraphics[width=\linewidth]{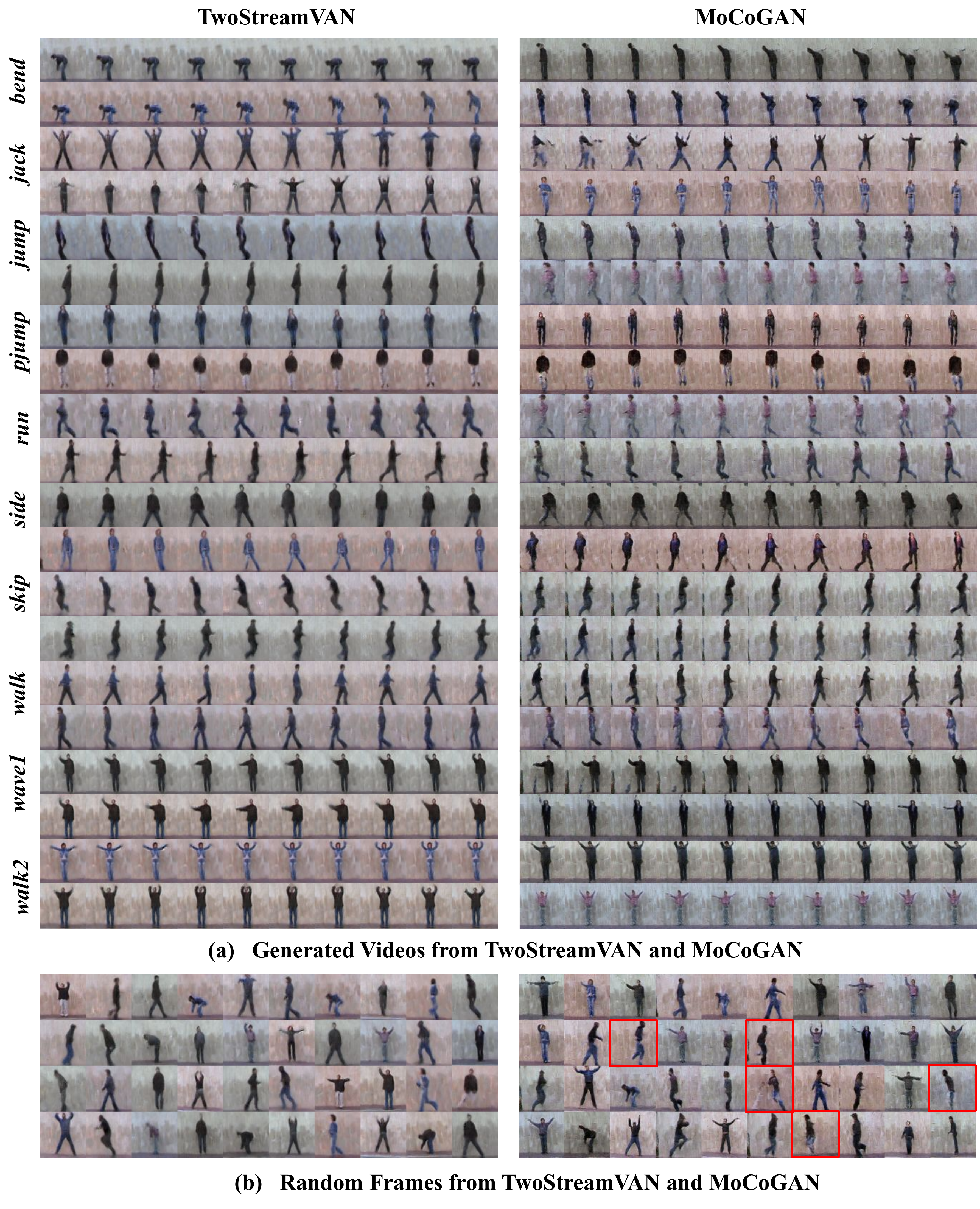}
    \caption{We provide 2 videos of one action class generated by TwoStreamVAN and MoCoGAN respectively on Weizmann Human Action Dataset. Also, we randomly sample 40 frames to show the content quality. We mark frames with large distortion in red.}
    \label{fig:weizmann_supplement}
\end{figure*}

\begin{figure*}
    \centering
    \includegraphics[width=\linewidth]{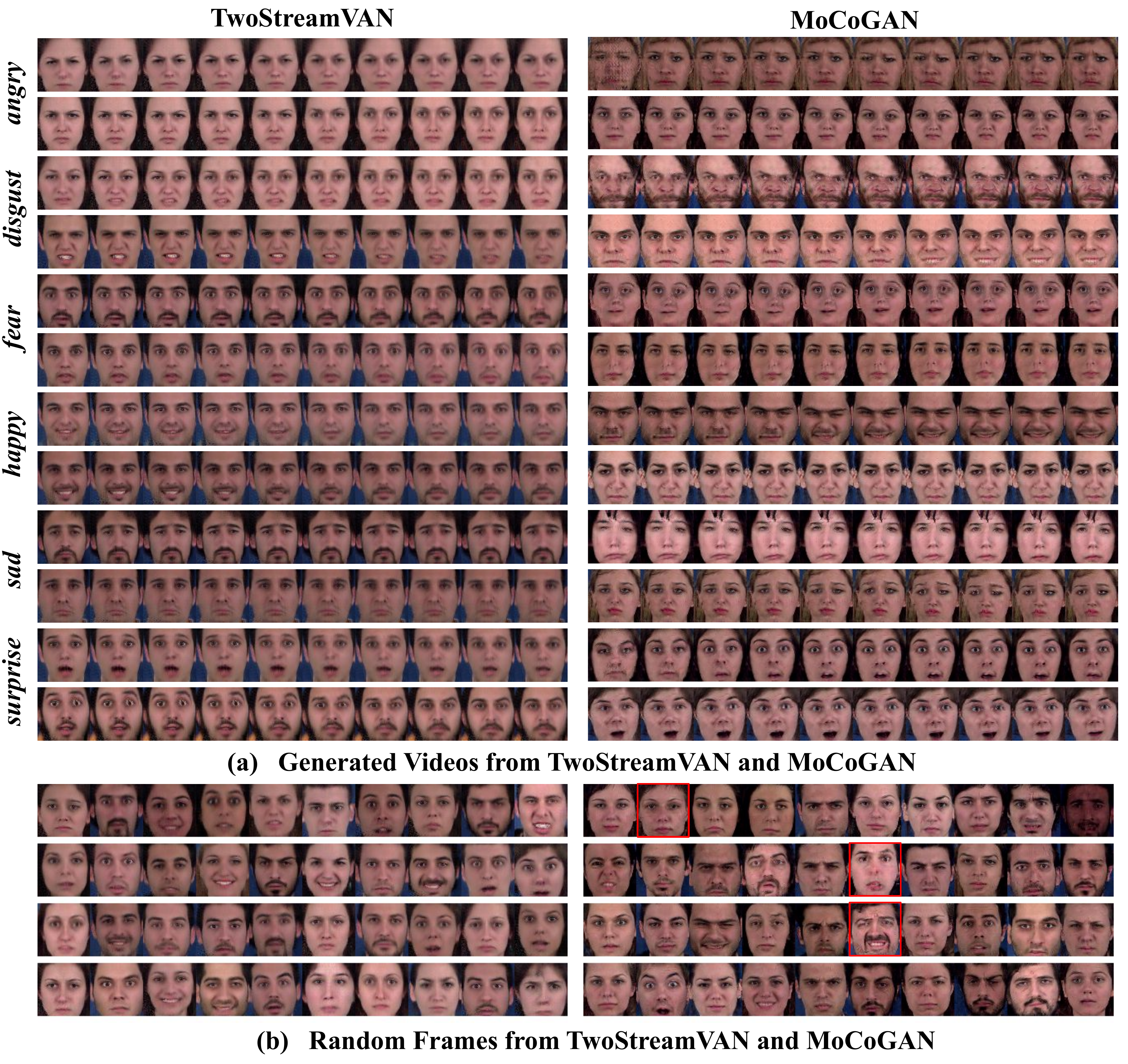}
    \caption{We provide 2 videos of one action class generated by TwoStreamVAN and MoCoGAN respectively on MUG Facial Expression. Also, we randomly sample 40 frames to show the content quality. We mark frames with large distortion in red.}
    \label{fig:mug_supplement}
\end{figure*}

\begin{figure*}
    \centering
    \includegraphics[width=\linewidth]{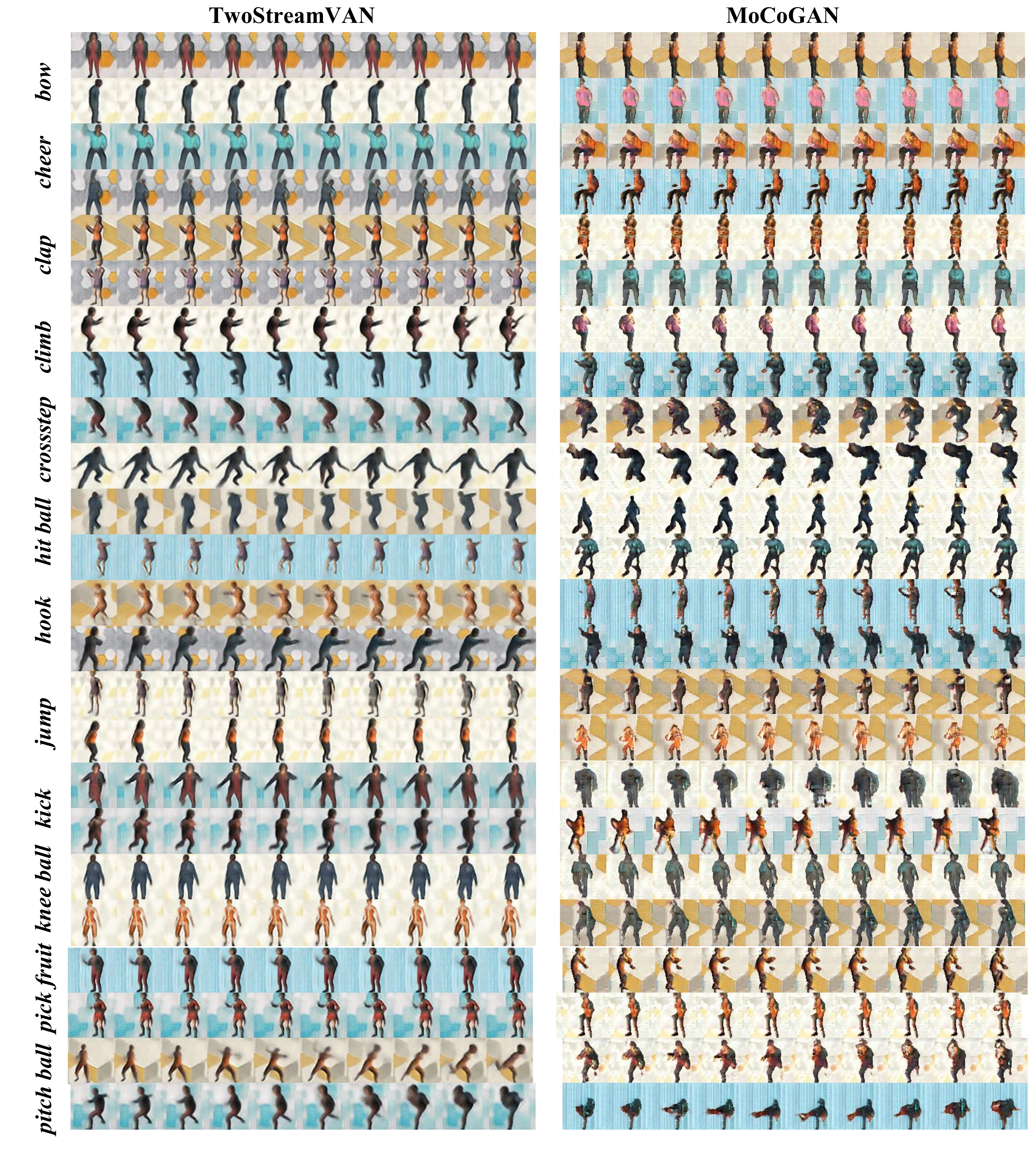}
    \caption{We provide 2 videos of one action class generated by TwoStreamVAN and MoCoGAN respectively on the first 12 classes of SynAction.}
    \label{fig:synaction_supplement1}
\end{figure*}

\begin{figure*}
    \centering
    \includegraphics[width=\linewidth]{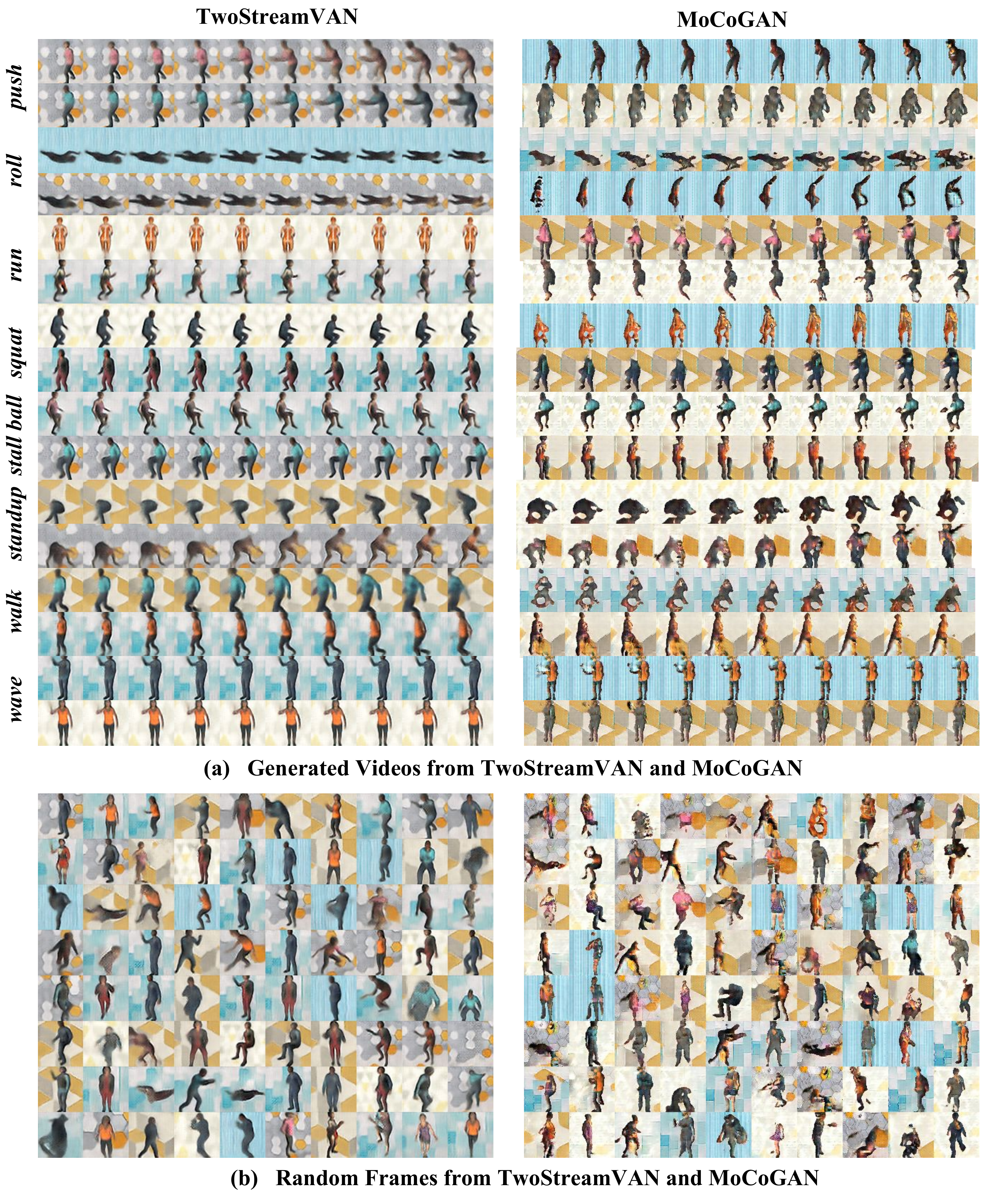}
    \caption{We provide 2 videos of one action class generated by TwoStreamVAN and MoCoGAN respectively on the remaining 8 classes of SynAction. Also, we randomly sample 80 frames to show the content quality.}
    \label{fig:synaction_supplement2}
\end{figure*}

\begin{figure*}
    \centering
    \includegraphics[width=\linewidth]{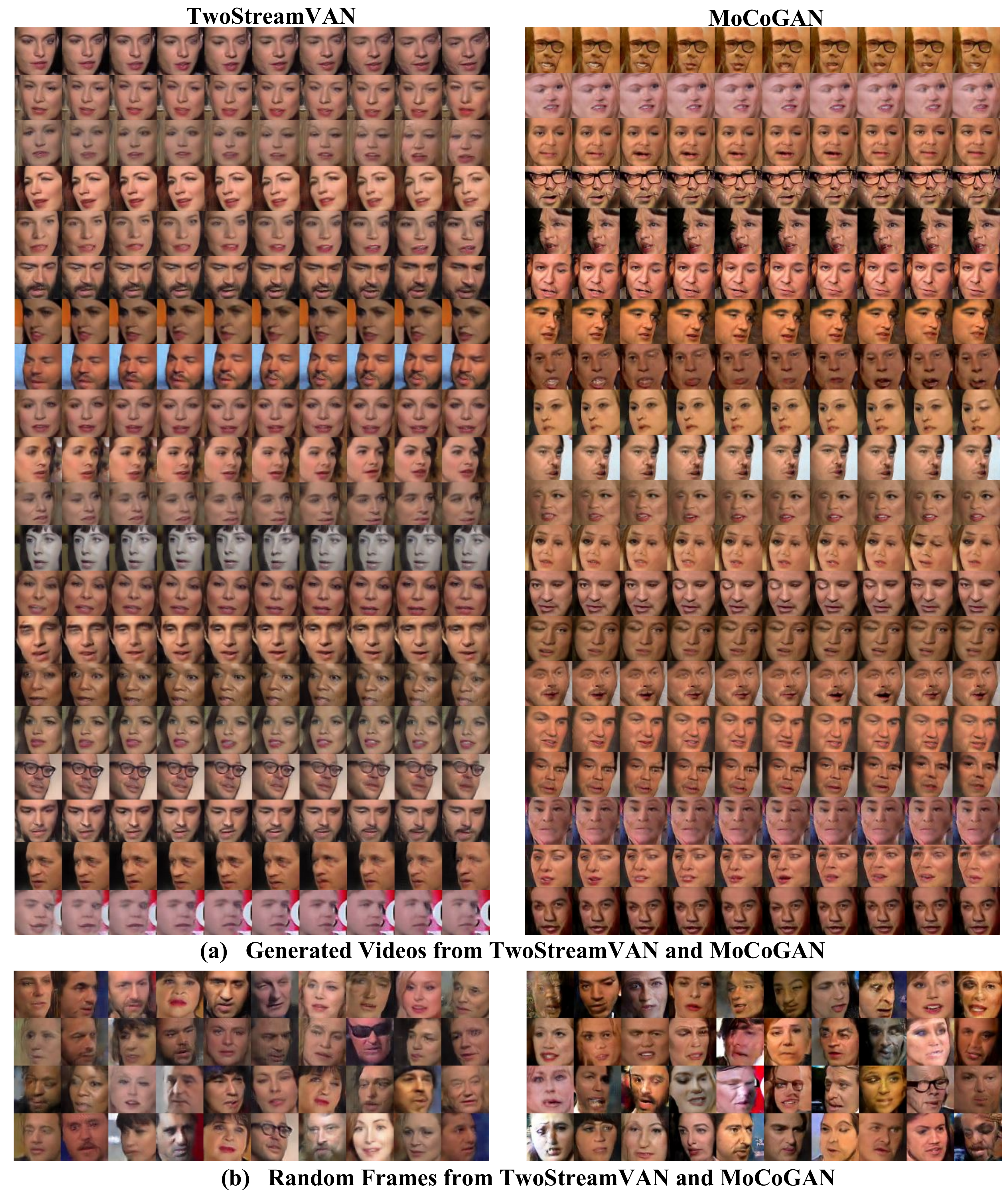}
    \caption{We provide 20 videos generated by TwoStreamVAN and MoCoGAN respectively on VoxCeleb. Also, we randomly sample 40 frames to show the content quality.}
    \label{fig:voxceleb_supplement}
\end{figure*}

In this section, we provide more qualitative visualizations of generated videos from TwoStreamVAN and MoCoGAN~\cite{tulyakov2017mocogan} on each dataset (Fig.~\ref{fig:weizmann_supplement} for Weizmann Human Action~\cite{ActionsAsSpaceTimeShapes_pami07}, Fig.~\ref{fig:mug_supplement} for MUG Facial Expression~\cite{aifanti2010mug}, Fig.~\ref{fig:synaction_supplement1} \& \ref{fig:synaction_supplement2} for our SynAction Dataset and Fig.~\ref{fig:voxceleb_supplement} for VoxCeleb~\cite{Nagrani17}). On Weizmann, MUG and SynAction Datasets, We provide 2 videos as examples for each action class: videos from TwoStreamVAN and MoCoGAN are generated with the given class. On VoxCeleb, we provide 20 videos from TwoStreamVAN and MoCoGAN respectively. Due to the fine-grained motion included, we recommend readers to view the video version of this visual comparison in the supplementary material.

\end{document}